# CLiFF Notes

### Research in the
## Language, Information and Computation Laboratory
## of the University of Pennsylvania

Annual Report: 1994, No. 4

Technical Report MS-CIS-95-07
LINC Lab 282

Department of Computer and Information Science
University of Pennsylvania
Philadelphia, PA 19104-6389

Editors: Matthew Stone & Libby Levison

i

# Contents

## III    Projects and Working Groups    93

# Part I
# Introduction

This report takes its name from the Computational Linguistics Feedback Forum (CLiFF), an informal discussion group for students and faculty. However the scope of the research covered in this report is broader than the title might suggest; this is the yearly report of the LINC Lab, the *Language, Information and Computation Laboratory* of the University of Pennsylvania.

It may at first be hard to see the threads that bind together the work presented here, work by faculty, graduate students and postdocs in the Computer Science and Linguistics Departments, and the Institute for Research in Cognitive Science. It includes prototypical Natural Language fields such as: Combinatorial Categorial Grammars, Tree Adjoining Grammars, syntactic parsing and the syntax-semantics interface; but it extends to statistical methods, plan inference, instruction understanding, intonation, causal reasoning, free word order languages, geometric reasoning, medical informatics, connectionism, and language acquisition.

Naturally, this introduction cannot spell out all the connections between these abstracts; we invite you to explore them on your own. In fact, with this issue it's easier than ever to do so: this document is accessible on the "information superhighway". Just call up

```
http://www.cis.upenn.edu/~cliff-group/94/cliffnotes.html
```

In addition, you can find many of the papers referenced in the CLiFF Notes on the net. Most can be obtained by following links from the authors' abstracts in the web version of this report.

The abstracts describe the researchers' many areas of investigation, explain their shared concerns, and present some interesting work in Cognitive Science. We hope its new online format makes the CLiFF Notes a more useful and interesting guide to Computational Linguistics activity at Penn.

## 1.1 Addresses

Contributors to this report can be contacted by writing to:

> *contributor's name*
> *department*
> University of Pennsylvania
> Philadelphia, PA 19104

In addition, a list of publications from the Computer Science department may be obtained by contacting:

> Technical Report Librarian
> Department of Computer and Information Science
> University of Pennsylvania
> 200 S. 33rd St.
> Philadelphia, PA 19104-6389
> (215)-898-3538
> betsy@central.cis.upenn.edu

## 1.2 Funding Agencies

This research is partially supported by: ARO including participation by the U.S. Army Research Laboratory (Aberdeen), Natick Laboratory, the Army Research Institute, the Institute for Simulation and Training, and NASA Ames Research Center; Air Force Office of Scientific Research; ARPA through General Electric Government Electronic Systems Division; AT&T Bell Labs; AT&T Fellowships; Bellcore; DMSO through the University of Iowa; General Electric Research Lab; IBM T.J. Watson Research Center; IBM Fellowships; National Defense Science and Engineering Graduate Fellowship in Computer Science; NSF, CISE, and Instrumentation and Laboratory Improvement Program Grant, SBE (Social, Behavioral, and Economic Sciences), EHR (Education & Human Resources Directorate); Nynex; Office of Naval Research; Texas Instruments; U.S. Air Force DEPTH contract through Hughes Missile Systems.

## Acknowledgements

Thanks to all the contributors.

*Comments about this report can be sent to:*

> Matthew Stone or Libby Levison
> Department of Computer and Information Science
> University of Pennsylvania
> 200 S. 33rd St.
> Philadelphia, PA 19104-6389
> matthew@linc.cis.upenn.edu or libby@linc.cis.upenn.edu

**Part II**

# Abstracts

**Breck Baldwin**

Department of Computer and Information Science
breck@linc.cis.upenn.edu

## Uniqueness, Salience and the Structure of Discourse

In spoken and written language, anaphora occurs when one phrase "points to" another, where "points to" means that the two phrases denote the same thing in one's mind. For example, consider the following quote from Kafka's *The Metamorphosis*:

> As Gregor Samsa awoke one morning from uneasy dreams he found himself transformed in his bed into a gigantic insect.

The proper noun *Gregor Samsa* is the antecedent to the anaphor *he*, and the relation between them is termed anaphora. A thorny problem for those interested in modeling human language capabilities on computers is determining what an anaphor (the pointing phrase) picks out as its antecedent (the pointed-to phrase). Much of the difficulty arises in picking the right antecedent when there are many to choose from.

Since anaphora mediates the interrelatedness of language, accurately modeling this phenomenon paves the way to improvements in information retrieval, natural language interfaces and topic identification. I have developed an approach that is particularly suitable for applications that require broad coverage and accuracy. My approach achieves broad coverage by applying simple and efficient algorithms to structure the prior discourse; it maintains accuracy by detecting the circumstances under which it is unlikely to make a good choice.

A typical way to find the antecedent of an anaphor is to impose a total order on the entities evoked by the prior discourse via some saliency metric, and then to take the descriptive content of the anaphor and search the saliency order for a matching antecedent Hobbs [1], Sidner [5], Luperfoy [4]. This approach has the flaw that for any two candidate antecedents, one is more salient than the other. But examples like;

> Earl and Ted were working together when ?he fell into the threshing machine.

indicate that Earl and Ted are equally available as antecedents, hence the awkwardness of the example. My approach notes this problem by imposing a partial order on the entities of the prior discourse, which allows Earl and Ted to be equally salient, and requiring that there be a unique antecedent.

These changes have greater import than addressing linguistic nit-picks. From an engineering perspective, the changes allow the algorithm to notice when it does not have enough information to find a unique antecedent, and then appropriate fall back strategies can be used. For instance in quoted speech, it is better to wait for more information to arrive which follows the anaphor in the text so that the speaker is identified, thereby resolving uses of *I, you* and *me*. This is particularly important when trying to find inferred antecedents because it is a mechanism that can control search through knowledge bases–although I am just beginning to explore the possibilities in this area.

The structuring portion of the theory is achieved by assigning a part of speech tags and identifying noun phrases with a simple regular expressions. Next, I assign a partial order to explicit and inferred entities using an amalgam of grammatical cues Hobbs [1], Lappin [3] which are triggered by regular expression matches. A notion of local topic is provided by Centering Theory Grosz, Joshi and Weinstein [2]. Initial results show that the algorithm gets the correct antecedent for pronouns on the order of 85-90%, possessive pronouns 75% and the results for full definite NPs are still unknown–all percentages are for ambiguous contexts, i.e. two or more semantically acceptable candidate antecedents in the current and prior clause.

# Tilman Becker

Postdoc, Institute for Research in Cognitive Science
tilman@unagi.cis.upenn.edu

## Compact representation of syntactic knowledge

In the framework of the XTAG system, I am currently implementing *metarules* as an extension to the TAG formalism which will allow for a considerably more compact representation of grammars.

Lexical rules, called metarules in the LTAG framework, are used to capture the similarities between different trees for a lexical entry. They can cover morphological as well as syntactic phenomena. As a practical benefit, metarules allow for a significantly shorter description of grammars for natural languages; also, they capture linguistic generalizations which cannot be expressed in the original TAG framework.

The XTAG grammar uses the concept of *tree-families* to group together all the elementary trees "that share the same subcategorization type" (LTAG techreport, 1990) [1]. For example, all clauses which are headed by a transitive verb are grouped together in one tree-family. This includes variations like, e.g., wh-questions, relative clauses, topicalized and passive sentences. These similarities express linguistic relations between phrases within one elementary structure – an advantage of the extended domain of locality of TAGs. The same variations of the basic declarative sentence occur in almost all other tree-families (e.g., the tree-family for the di-transitive verbs) as well. In the current implementation of the grammar, these trees are explicitly included in all tree-families. This not only increases the number of trees significantly but also – and more importantly – misses a linguistic generalization: the existence of rules which can be used to describe such variations as passive, topicalization etc. across the tree-families.

Formally, metarules are introduced as a set of rules that transform a given initial set of elementary trees into an expanded (but still finite[1]) set of elementary trees. Thus they can be looked at completely independently from processing considerations (either generation or parsing).[2] However, the existence of this rules can be exploited by practical systems as well as by linguistic and cognitive theories.

## Other current work

Other current work is motivated by data from German and includes computational as well as linguistic aspects.

### Parsing Free Word Order in Polynomial Time

In joint work with Owen Rambow, we have extended the CYK style parser for TAG defined by (Vijay, 1987) [5] to give a polynomial time parser for a large subset of the V-TAG[3] languages.

---

[1] If such rules can produce infinitely many new elementary structures then the formal properties (e.g. the generative power) of the formalism can change. This is discussed in detail in (Becker, 1994) [2].

[2] This is in fact the approach we are pursuing in the XTAG system. The metarules will be used to expand a core grammar to a full grammar which in turn is used by the parser.

[3] See (Rambow, 1994) [4].

**Free Word Order and the Derivational Generative Capacity**

In (Becker et al., 1992) [3], we introduced the notion of 'derivational generative capacity' (DGC) as a new instrument for classifying the generative capacity of Linear Context-Free Rewriting Systems (LCFRS). The notion of DGC which formalizes the relations between predicates and arguments was used in [3] to show that no LCFRS grammar is powerful enough to derive Long-Distance Scrambling, a free word order phenomena of German, Korean and other languages. We are currently working on an extension of DGC to other grammar formalisms, e.g. to CCG and HPSG.

**Parasitic Gaps**

In a recently started project, Bernhard Rohrbacher and I are working on an analysis of the syntactic properties of parasitic gaps in German. We are interested in this question for a variety of reasons, among which are the suitability of new versions of the TAG formalism and differences to English due to the availability of 'scrambling' and overt case marking in German.

**Betty J. Birner**

Postdoc, Institute for Research in Cognitive Science
betty@babel.ling.upenn.edu

## Discourse functions of syntactic constructions

Speakers may choose from among a variety of semantically equivalent ways of expressing a given proposition; what syntactic construction is used in a given context is governed by discourse-level considerations. Inversion, topicalization, and *there*-insertion in English have all been found to depend for their felicity on related yet distinct factors involving the discourse status of the information represented within the utterance; these findings are supported by the distribution of these constructions in naturally-occurring discourse. Based on these data, generalizations can be drawn concerning what pragmatic characteristics are shared by related constructions, and what characteristics are unique to particular constructions. I am currently working in collaboration with Gregory Ward of Northwestern University on the functions of leftward movement in English, addressing the functional distinctions between canonical word order and fronting constructions such as inversion and topicalization as well as the various types of links that may relate these utterances to the prior discourse. In related work, we have found that the so-called 'definiteness effect' associated with *there*-sentences in English is not categorical, but rather reflects an interaction of pragmatic constraints on *there*-sentences and definiteness; these results in turn led us to develop an account of the felicitous use of the definite article in English.

I am also conducting research in collaboration with Shahrzad Mahootian of Northeastern Illinois University into the function of inversion in Farsi, as a first step in determining whether cross-linguistic generalizations may be drawn regarding the function of inversion. Farsi is an SXV language, and permits marked orderings of either XVS (linearly like English inversion) or XSV (linearly like English topicalization). We hypothesize, however, that the XSV ordering in Farsi corresponds functionally to English inversion. This finding would support the notion that constructions are functionally defined, and that discourse function is not tied to syntactic structure cross-linguistically (Prince, 1988) [8].

Finally, my major research project involves examining the nature and status of inferrable information in discourse. In both topicalization and inversion in English, inferrable information appears to be treated as 'discourse-old' (Prince 1992) [9], i.e. familiar within the discourse; one of my goals for the Farsi study is to determine whether this generalization holds as well cross-linguistically. These questions have ramifications for the representation of inferrable information within the discourse model. A related issue concerns the relationship between intonation and inferrable information; although inferrables in inversions behave as discourse-old for purposes of word order, they behave as discourse-new intonationally. Using data from a variety of constructions and languages, I hope to formulate a unified account of the linguistic representation of inferrable information that will shed light on both its treatment in discourse and its intonational realization.

# Sandra Carberry

Visitor, IRCS and Department of Computer and Information Science
carberry@linc.cis.upenn.edu

## User Modeling in Patient-Centered Activity

Keywords: Information delivery, User modeling, Medical informatics

During the past year, I have been working with Bonnie Webber on TraumAID, a decision-support system aimed at improving trauma care during the initial definitive phase of patient management. I am interested in how a system can most effectively interact with members of a trauma team in order to have a beneficial impact on patient care. Potential intervention modes vary primarily according to content (full management plan or critique of the physician's decision-making), person with whom the system interacts, and presentation modality (text, speech, graphics). I have been analyzing videotapes of trauma care to identify how information is communicated among members of a trauma team and how a system might effectively intervene.

I am particularly interested in exploring how the effectiveness of the system's communication might be increased via user modeling. Systems that employ user modeling exploit a model of an agent's knowledge, beliefs, abilities, goals, etc. to tailor the system's actions to individual agents. While little attention has been given to user modeling in medical decision support systems, it has been validated in other areas. Our hypothesis is that tailoring messages to the credentials of the message recipient can increase acceptance and utilization of the system's input. We are initially considering two forms of message tailoring: the way the message is framed and the level of justification accompanying it. Varying *how* the message is framed consists of changing the illocutionary force of the communicative goal, perhaps from a suggestion to a reminder, thereby altering the structure and phrasing of the message. Since reminding carries the presupposition that the hearer knows the communicated information but may have overlooked it, the system might *remind* an experienced trauma surgeon in instances where it would be appropriate to suggest a procedure to a trauma surgery resident. Varying the justification included with a recommendation consists of varying the evidence explicitly given for adopting a belief or undertaking an action. Since different message recipients have different detailed medical knowledge, tailoring the justification to the credentials of the recipient should increase the utility of the system's communication.

The TraumAID knowledge base contains a wealth of diagnostic and therapeutic information that might also be utilized for other purposes, such as educating medical personnel. If a knowledge base such as TraumAID's could be used to construct cases dynamically, an intelligent tutoring system would not need to maintain a library of existing cases. Moreover, the system could tailor each generated case to the student's current expertise and modify it as the student works on the case and exhibits reasoning deficiencies. In conjunction with Abigail Gertner, I have begun to investigate the knowledge that must be added to TraumAID in order to construct *interesting* cases and a strategy for generating cases using the TraumAID knowledge base.

## Participating in Collaborative Expert-Consultation Dialogues

Keywords: Dialogue, Plan inference, Collaboration

For several years I have been interested in collaborative consultation dialogues in which an executing agent and a consultant are working together to construct a plan for achieving the executing agent's domain goal. Examples of such goals include making financial investments, purchasing a home, or obtaining a university degree. The executing agent and the consultant bring to the plan construction task different knowledge about the domain and the desirable characteristics of the resulting domain plan. For example, the consultant presumably has more extensive knowledge of the domain than does the executing agent, but the executing agent has knowledge about his particular circumstances, intentions and preferences that are either restrictions on or potential influencers [1] of the domain plan being constructed. So the two participants play different roles in the dialogue.

A crucial component of a collaborative consultation system is the system's beliefs about the user's intentions as communicated during the dialogue. These can be domain intentions (such as making an investment), problem-solving intentions (such as comparing alternative ways of performing an action), or communicative intentions (such as answering a question or expressing doubt at information conveyed by the other participant). Previous work [6] resulted in a tripartite model that distinguishes among domain, problem-solving, and communicative actions yet captures the relationships among them. Recently, our work has taken several directions, including:

- the development of a plan-based response generation component [2],

- a computational strategy for generating [4] and interpreting [5] indirect replies, and

- a strategy for dynamically recognizing user preferences during the course of a collaborative consultation dialogue [3].

# Christine Doran

Department of Linguistics
cdoran@linc.cis.upenn.edu

## Bootstrapping a CCG from XTAG

Keywords: Categorial Grammars, Tree Adjoining Grammars, Parsing

Currently, a number of small-scale CCGs and parsers exist, but, to our knowledge, there have been no attempts to construct a large-scale CCG parser with the lexicon to support it. On the other hand, the XTAG project has an implemented wide-coverage lexicalized TAG (LTAG) system. The formal and structural similarities between LTAGs and CCGs [4] [5] are such that we felt we could use the XTAG system to bootstrap a CCG[1] [1].

We have created a mapping between LTAG trees and CCG categories, based on the 200 trees used in an XTAG-parsed corpus of 6,000 Wall Street Journal sentences. In addition to translating the trees, we mapped the grammatical features from the trees to the CCG categories. The next step, currently in progress, is to use the category/tree mapping to build the CCG syntactic and category databases from their XTAG counterparts. The translations will also be used to collect statistics on category frequency from the the XTAG-parsed corpus, for use in bootstrapping the statistical filters. Since the mapping is not one-to-one, the statistics will not be precise, but will nonetheless serve to speed up the CCG parser and enable us to collect more accurate statistics on actual CCG derivations.

Thus far, the most interesting grammatical issue in translating LTAG to CCG is that, while they have the same formal power and superficially similar representations, certain parts of the grammatical workload are distributed differently. The two primary differences we found in translating the XTAG grammar to CCG involve the handling of extraction – whole sets of XTAG trees collapse into single CCG categories – and the the accessibility of VPs and non-constituents in CCG which are not accessible in XTAG.

See Al Kim's abstract in this volume for information on the parser.

## Work for the XTAG Project

Keywords: Syntax, Parsing

### Heuristics for Ranking Parses

The XTAG system [1] consists of a wide-coverage parser, supported by an extensive grammar and lexicon; while the parser uses some statistical information for part-of-speech and "supertag" disambiguation, it does not have the corpus-specific statistical knowledge or the semantic knowledge necessary to choose among ambiguous parses. As a result, there are often a number of parses for a given sentence (the average is 6, for sentences of 15 or fewer words). As part of the disambiguation process for XTAG, we have developed a set of heuristics for ranking parses using structural information about individual trees and about the derivation trees [7]. These heuristics work with the two-stage parsing process and with the existing statistical filters to produce a ranked list of parses.

The disambiguation works as follows. Part-of-speech (POS) tagging is first used to minimize the number of trees selected. Following that, supertagging is used to select the three most likely trees for each word; certain of our heuristics apply to this winnowed set of trees, ranking some constructions across the board (e.g. disprefer topicalization) and some for specific lexical items (e.g. prefer *be* as an auxiliary verb). The remaining trees are sent to the parser, which generates parses in a rank order. This ranking is determined using a combination of the weights obtained prior to parsing and weightings based on the derivation trees of the generated parses (e.g. for right- vs. left-branching structures). For the sentence *Each pool should have an activity level of one and only one*, this combined disambiguation strategy reduces the number of parses from an unranked list of 60 to a ranked list of 3.

---

[1]Work done with B. Srinivas, Julie Bourne, Hoa Dang, David Fine, Al Kim and Mark Steedman.

This ranked list can be used to control the number of sentences passed on to further levels of processing. In applications emphasizing speed or in evaluations of the parser against a gold standard, only the highest ranked parse would be considered. In applications emphasizing accuracy, the top $N$ parses can be considered. Similar heuristics have been used for other parsers (e.g. recent work by Hobbs and Bear [3] and McCord [6]).

### Analysis of Embedded Sentences

The XTAG analysis of sentential complements and sentential subjects has been extended, and we have added an analysis of sentential adjuncts. A successful analysis of complementizers in English must address both the co-occurrence restrictions between complementizers and various types of clauses, and the distribution of the clauses themselves, in both subject and complement positions. The XTAG grammar now meets these criteria, via features on both the complementizers and the matrix verbs. Thus, sentences such as *The dog wants whether to go out* and *The cat insisted to go out* are blocked.

A similar analysis applies both to sentential adjuncts, such as purpose clauses (1), and to clauses coordinated with subordinating conjunctions (2):

1. To impress Bill, Max shaved his head.

2. Because he wanted a new look, Max shaved his head.

# Dania Egedi

Research Programmer, Institute for Research in Cognitive Science
egedi@ahwenasa.cis.upenn.edu

## XTAG Project Related Work

### Determiner Ordering

This is joint work with Beth Ann Hockey.

One question that has not been extensively discussed in previous work on determiners is how the class of determiners order with respect to each other. We have identified a set of determiner features motivated solely by semantic properties that when used together have the consequence of accounting for the complex patterns of syntactic determiner sequencing. Although we do not claim to have exhaustively covered the rich determiner system of English, we do cover a large subset. This analysis is fully implemented as part of the XTAG project.

In our account of determiner sequencing [3], there is only one class of determiners. Each determiner carries with it a set of features that represents its properties, and a set of features that represents the properties of any determiners it may modify. This account encompasses other determiner constructions as well, such as numbers, genitive and partitive constructions. We are currently working on incorporating adverbial modifiers of determiners.

### FTP'able Syntactic and Morphological Databases

With the advent of Machine Readable Dictionaries (MRDs), the problems of creating large-scale lexicons changed from the tiresome, painstaking task of trying to develop individual word lists for various syntactic phenomena to the task of 'simply' extracting the information from the on-line dictionaries. MRDs, however, present numerous problems in terms of the errors and inconsistencies in the various components of the entries, making extraction quite difficult. In the interest of sharing resources, the XTAG project is making available the syntactic and morphological databases developed for the XTAG wide-coverage grammar.

The **morphology database** [4] was originally extracted from 1979 edition of the Collins English Dictionary and Oxford Advanced Learner's Dictionary of Current English, and then amended and augmented by hand. It consists of approximately 317,000 inflected items, along with their root forms and inflectional information (such as case, number, tense). Thirteen parts of speech are differentiated: Noun, Proper Noun, Pronoun, Verb, Verb Particle, Adverb, Adjective, Preposition, Complementizer, Determiner, Conjunction and Interjection, and Noun/Verb Contraction. Nouns and Verbs are the largest categories, with approximately 213,000 and 46,500 inflected forms, respectively. The access time for a given inflected entry is .6 msec.

The **syntactic database** [2] associates lexical items with the subcategorization frames based on selectional information. The syntactic database entries were originally extracted from the Oxford Advanced Learner's Dictionary and Oxford Dictionary for Contemporary Idiomatic English, and then modified and augmented by hand. There are more than 37,000 syntactic database entries. Each entry consists of an INDEX field, the uninflected form under which the lexical item is compiled in the database; an ENTRY field, which contains all of the lexical items associated with the INDEX; a POS field, which gives the part-of-speech for the lexical item(s) in the ENTRY field; and then a FRAME field, which contains the syntactic information about that entry. An optional FS (feature structure) field may provide additional information about the FRAME field. Any number of EX fields may be provided for example sentences. Lexical items may (and usually do) have more than one entry in the database and they may select the same FRAME field more than once, using the feature structures (FS) to capture lexical idiosyncrasies.

Both databases are in a hash table format, and each comes with its own X-window-based interface program for use in customizing the databases, as well as C and Lisp hooks for use in other programs. The databases and interface programs were all developed on the Sun SPARC station series. More information and ftp instructions can be obtained by sending mail to lex-request@linc.cis.upenn.edu.

## Learning New Concepts through Metaphors

Keywords: Metaphor, Learning, Computational, Semantic nets

Humans often explain new concepts by relating them to already known, well-understood concepts. We set up an analogy between the well-understood and the new concept and then we are able to talk about the new one using the terminology associated with the old. In this way, we come to understand exactly where the metaphor applies and where it does not. What we end up with is a useful and accessible concept that can stand on its own without the intermediate one that was used to explain it.

I am continuing work on MLS (the Metaphor Learning System) [1], a computational system that takes advantage of these type of metaphors to expand its knowledge-base. MLS is given a number of sentences that use the unknown target concept as though it were the known source concept (i.e. *Electricity flows through the wire*). For each sentence, the systems enters into the knowledge representation as though it were the source concept, and then abstracts back up the hierarchy until it finds a place where the constraints on the target concept match. This process is called **abstraction**. Once the proper place is found, the system places the new information into the knowledge-base by a process known as **concretion**. Links are tenuous, and become stronger or weaker depending on subsequent sentences. Repeated exposure to a number of sentences that use the target concept metaphorically allows the system to build a representation of the target concept that can then be used for other purposes, including serving as the source concept for another concept to be learned.

# Jason Eisner

Department of Computer and Information Science
jeisner@gradient.cis.upenn.edu

## Probabilistic Parsing

Parsing may be thought of as a search problem. Which meaning, from the space of all possible meanings, was a sentence intended to convey? A perfect theory of the language would rule out all but a few possible answers—but it might not make it easy to *find* those answers. In any case, present-day NL systems do not have perfect syntactic theories. Nor do human children.

Probabilistic parsers guide their search with a few hard constraints and many soft constraints. The hard constraints come from Universal Grammar or a grammar of the target language. (They might insist that constituents be contiguous, or that prepositions subcategorize for noun phrases.) The soft constraints compensate for the paucity of hard constraints by capturing syntactic and semantic *preferences* of the language. The observable tendency of "water" to be a noun rather than a verb, and its preference for plant objects when it *is* a verb, are soft constraints that can nudge a parser toward the correct parse. Such preferences can be described in probabilistic terms. Their strengths can be empirically estimated from a "training corpus" of (sentence, interpretation) pairs.

A probabilistic parser that I recently built can annotate natural text with part-of-speech tags on the words and directed, unlabeled dependency links among the words. It uses a new cubic-time dynamic programming algorithm to find the most probable global joint assignment of tags and links. There is no grammar: the only hard constraint is that the result be a valid dependency grammar parse (a tree without any crossing branches).

The potential advantage of the dependency grammar framework is that all parsing decisions are lexical. All the "natural" probability measures on parses are sensitive to lexical preferences—a bonus when the parser is well-trained. However, the parser has performed promisingly even on the tiny amounts of training it has received so far. It can assign two-thirds of the links correctly, at no overall cost or benefit to an 81% tagging success rate, even when a whopping 60% of the words are unknown. (The system incorporates a novel, principled approach to the unknown-words problem.)

It is entirely plausible that humans use soft syntactic and semantic constraints to help them understand language accurately and efficiently. Unlike the parser above, however, humans appear to do their processing from left to right in linear time. They seem to prune unlikely partial parses before right context becomes available. I hope to resume some past research on probabilistic parsers having these broad characteristics. Early versions of that work, conducted with Mark A. Jones, are described in [2] and elsewhere.

## Normal-Form Parsing in CCGs

Combinatory categorial grammars (CCGs) provide an intriguing model of grammatical competence. However, they have the irksome habit of assigning a great many parses to even an unambiguous sentence. All these parses are semantically indistinguishable: any one of them would be adequate to represent the entire set.

Such redundancy takes its toll on CCG parsers. Finding all the permissible parses takes time; finding the few distinct meanings among the many parses also takes time. Moreover, the indifference of CCG to some syntactic choices would make it difficult to construct or train a probabilistic CCG parser.

Fortunately, these problems can be overcome. I have shown that it is possible to build an efficient parser that constructs a single canonical member of each semantic equivalence class. The result holds for CCG grammars that include (arbitrarily restricted) higher-order composition and crossed composition, as described in Joshi *et al.* [3]. I do not yet have an analogous result for grammars that permit type-raising or crossed substitution [5].

## A Unified Semantics for English "Any"

Keywords: Any, Negative polarity items, Quantification

English *any* is often treated as two unrelated lexemes: a existential quantifier that may appear in the scope of negation ("We don't have any bananas"), and a universal quantifier ("We can order any bananas we don't have").

Influential early papers on *any* [1] [4] presented arguments for this distinction. I propose that there are good reasons for abandoning it, and treating *any* uniformly as a universal quantifier. If both senses are taken to be universals, they have the same idiosyncratic scope properties. Unifying them also permits a natural account of the environments where *any* is allowed. Finally, the hypothesis that *any* raises over negation is supported by relative scope tests and other data.

# Christopher Geib

Department of Computer and Information Science
geib@linc.cis.upenn.edu

## Intentional Action and Planning

Keywords: Planning, Intentionality, Actions, Effects

My research is about the design of an efficient planning system that recognizes the context dependent nature of the effects of actions and the effect of an agent's intentions on action choices. Consider the action of an agent opening its hand while holding an object. Suppose that the agent performs this action in two different situations. In the first situation, the object in the agent's hand is supported by some other object. In the second situation, the object is not supported. If the agent opens its hand in the first situation, the first object will be stacked on the second. However, if the agent performs the action in the second situation, the object will fall and possibly break. Thus, because different conditions hold at the time of the execution of the motion different actions are produced by its performance, namely the stacking action and the dropping action.

Existing planning systems [4], [5] suffer from the following problem. Suppose that the system has an action for accomplishing the first action. That is, they have a release action that is used to stack objects on other objects. If they wish to use the action of opening the hand to drop the object and break it, they are required to create a new, separate action to achieve this result. This has two unfortunate side effects. First it increases the size of the search space that the system must explore. Second this solution introduces redundancy in the knowledge base of the agent. This solution will solve the problem but at the expense of multiplying the runtime of the planning algorithm by a larger number of actions. Finally it does not conform to our intuitions that there is something that is the same between these two actions, namely the opening of the hand, that is not captured by the creation of the new action.

Part of this problem is a result of an inability of many systems to distinguish between an action and its effects [3], [4]. I take it as straightforward that *actions* are sets of movements that an agent performs while *effects* are the concrete changes in the world that are a result of actions. While there is a causal relationship between an action and its effects it is important to remember that this relation is not universal. An action can be taken without a given effect and a given effect can be caused by a number of actions. In the example above, the effect in one case is that the object is stacked on top of another object, and in the other that the object has fallen and is now broken.

The goal of my work is to provide an efficient planning system that will allow the use of actions for any of their effects without introducing the overhead of redundant actions. I believe that this requires a fundamental reconsideration of the roles that actions, effects, and preconditions play in the planning process and their relationship to goals and intentions. I have implemented these ideas in a planner called ItPlanS (the Intentional Planning System) and this planner has been successfully used in three domains, a two handed blocks world [1], the SodaJack project [2], and a short order cooking domain.

# Abigail S. Gertner

Department of Computer and Information Science
agertner@linc.cis.upenn.edu

## Critiquing Trauma Management Plans On-Line

Keywords: Decision support, User interfaces, Plan recognition, Plan evaluation

Decision-support systems have typically been designed so that the interaction between user and system takes place during an "off-line" consultation session. The user is assumed to be sitting in front of the computer terminal, either engaged in some activity which the system can directly monitor to provide help when necessary, or for the specific purpose of consulting with the system. Under these circumstances, the user's attention is focused primarily on the output of the system, which can be quite detailed in its explanation and discussion.

However, in certain domains it is not practical for a decision-maker to sit down in front of a computer terminal in order to validate his plan prior to carrying it out. Instead, information delivery in these situations must be performed while the the user's attention is focused elsewhere than on the system. This requirement is the consequence of two domain features: first, the activity with which the user is involved requires constant attention while it is in progress, and second, the amount of time available for executing actions is limited.

I am working on an approach to communicating information from system to user that can be described as *task-centered* rather than *system-centered*. In this approach, the goal of the system is to present critical information in a succinct and timely manner in order to maximize the salience of that information to the user. Information that is relevant but is not considered critical to the outcome of the task is not included in the output of the system. I have adopted a *critiquing* approach to decision support (see [3]), in which the output of the system is in the form of a critique based on the user's intended actions rather than the more traditional expert system approach of providing a recommended plan to follow. Critiquing has the advantages that (1) it reinforces the role of the user as primary decision maker, and (2) since the critique is based on the user's intended actions it provides a user-oriented focus for its output.

This work has been implemented in TraumaTIQ [1], [2], the output module for TraumAID, a decision-support system for the delivery of trauma care during the *initial definitive* phase of patient management [5], [6], [7]. TraumaTIQ is responsible for interpreting proposed actions in the context of the current state of the patient, and producing a critique of those actions.

The critiquing process in TraumaTIQ is triggered when either the physician orders some action or actions to be performed, or new information about the patient – such as findings, test results, actions performed, etc. – is entered into the system, in which case the physician's plan will be reevaluated in light of that new information. Orders for actions are taken to reflect the intention to perform those actions and thus serve as the object of the critique.

The architecture of TraumaTIQ comprises three components: First, plan recognition allows the system to infer the likely goals underlying the physician's proposed actions in the context of the patient's current condition. Second, the plan evaluation phase identifies discrepancies between the physician's plan and TraumAID's, and determines which of these discrepancies are critical enough to be mentioned in the critique. Finally, critique generation translates the comments identified by the plan evaluation stage into English sentences to be output to the physician.

By comparing the model of the physician's plan with the plan developed by TraumAID, TraumaTIQ is capable of recognizing four different types of discrepancies:

- Omission: in which the physician appears not to be addressing a goal in a timely manner.

- Commission: in which the physician has ordered a procedure that the system does not consider motivated in the current situation.

- Procedure Choice: in which the physician has ordered a procedure to address a valid goal, but that procedure is not the one preferred by TraumAID to address that goal.

- Scheduling: in which the physician addresses a less important goal before a more important one.

These discrepancies are then evaluated according to whether they are (1) tolerable, (2) non-critical but potentially harmful or (3) critical. This judgement is based on the type of error, the costs associated with it, and/or the risk associated with not doing it (in the case of an error of omission). Only items in the second or third category will be mentioned in the critique. Finally, comments are translated into English sentences using templates corresponding to the different types of errors the system recognizes. Future investigations will examine the relative effectiveness of different methods of conveying comments to the physician: (1) displaying them directly on a monitor in the physician's view, (2) having a nurse read them out loud, (3) generating them with synthesized speech using a system such as the one described in [4].

**James Henderson**

Department of Computer and Information Science
henders@linc.cis.upenn.edu

## Connectionist Syntactic Parsing Using Temporal Synchrony Variable Binding

The ability to learn to combine multiple sources of soft constraints has made connectionist networks important tools. On the other hand, their inability to dynamically manipulate complex compositional representations has prevented them from being successfully applied to many problems for which traditional symbolic methods have proven important. Natural language parsing requires both these abilities. Recent work on how to support symbolic computation within a biologically motivated connectionist computational architecture (that of Shastri & Ajjanagadde (S&A) [3]) has made the combination of these abilities possible, but the S&A architecture has some limitations. My research investigates the implications of the computational constraints imposed by the S&A architecture for natural language syntactic parsing. My dissertation presents a model of syntactic parsing in this architecture, and uses it to demonstrate that these constraints do not pose problems for syntactic parsing. In addition, these computational constraints are argued to make interesting predictions about the nature of language, specifically in the areas of long distance dependencies and center embedding. By showing how syntactic parsing can be done in the S&A connectionist architecture, this work allows the advantages of connectionist approaches to be applied to natural language parsing, while still taking advantage of symbolic approaches, and making use of the large body of existing work on natural language.

The inadequacies of previous connectionist parsers can be attributed to the inability of their architectures to separate information about what constituents are in the phrase structure of the sentence from information about what features those constituents have. In symbolic representations, variables are used to specify the former information, and predicates the later. A connectionist model of computation recently proposed by Shastri and Ajjanagadde uses the temporal synchrony of activation to represent the bindings between different features of the same variable (i.e. temporal synchrony variable binding). With this added temporal dimension for representing information, the S&A architecture can store and dynamically manipulate predications over variables, thus supporting symbolic computation. In addition, it is biologically motivated, it allows the massively parallel use of knowledge, and it supports evidential reasoning. Although this architecture eliminates the core problem encountered in previous investigations of connectionist parsing, it does have some limitations. It has a bounded memory capacity, it can only store a conjunction of predications, and in the general case it is costly to store and process relations between variables.

The limitations of the S&A connectionist architecture can be solved through the use of partial descriptions of phrase structure trees. Because a parser should produce incrementally interpretable output and the architecture can only store a conjunction of predications, the parser's grammatical representation needs to be sufficiently partial to allow it to state all and only what it knows in a conjunction of predications. My masters thesis [1] proposes a grammar formalism, called Structure Unification Grammar (SUG), which is designed to comply with this requirement. SUG is a formalization of accumulating partial information about the phrase structure of a sentence until a complete description of the sentence's phrase structure tree is constructed. My masters thesis demonstrates that SUG is a powerful, flexible, and perspicuous grammatical framework by showing how analyses and insights from a variety of other grammatical investigations can be captured using SUG. Partial descriptions are also required in order to parse in bounded memory, since the parser must abstract away from information which it no longer needs. SUG also supports an operation which allows unneeded constituents to be forgotten without risking the violation of the forgotten constraints.

My dissertation work uses Structure Unification Grammar as a grammatical framework in which to investigate efficient syntactic parsing in the S&A connectionist computational architecture. The architecture's bounded memory capacity is handled by incrementally outputting information about the phrase structure of the sentence, and forgetting phrase structure nodes for which no new information will be added. In this way the parser can parse arbitrarily long sentences without running out of memory. The partiality of SUG descriptions allows the parser to state only the information it is sure of (as is done in Description Theory [2]), rather than stating a disjunction of more completely specified alternatives. This allows the parser to do disambiguation incrementally without needing disjunction. Rules which involve relations between variables can be eliminated through the use of a bounded stack. Relations can be accessed with respect to the unique node on top of the stack, thereby turning the binary relations into unary predicates.

My dissertation argues that the above techniques allow a parser implemented in the S&A connectionist architecture to be adequate for recovering the syntactic constituent structure of natural language sentences. This argument is given in the form of an existence proof, presenting a specific parsing model and specific grammatical analyses. This parser has been tested in all the areas which are of particular concern given the limitations of the architecture, specifically phrase structure analyses, long distance dependencies, center embedding, and representing local ambiguities. It has also been tested on a set of sentences randomly selected from the Brown corpus. None of this testing uncovered any phenomena which would be harder for this architecture than they are in general. In addition to this argument for the adequacy of the architecture, it is argued that the constraints of the architecture make significant predictions about the nature of language. These predictions are mostly in the areas of long distance dependencies and center embedding, and are largely due to the bounded stack introduced for processing binary relations.

## Beth Ann Hockey

Department of Linguistics
beth@linc.cis.upenn.edu

## The Interpretation and Realization of Focus

The term *focus* has been used by a variety of authors within linguistics to mean a very wide variety of things. Some authors use the term "focus" for prosodic prominence, others for referential status, information structure, contrast and association with particles like *even* and *only*. While these notions may have similarities, be related to or interact with each other, they are not obviously the same object. It is an open question whether there is a single overarching notion of focus on which a unified treatment of the many phenomena can be based, or whether distinguishing among a number of types of focus will make more sense of the puzzle. The goal of this work is to develop interpretation–realization pairing(s) for notion(s) of focus that can make sense of the confusing and conflicting literature on focus and provide an explicit statement of the notions underlying focus and the relations between those notions.

## Echo questions, intonation and focus

English echo questions (EQ's) are not ordinary wh-questions (OQ's), but rather focus constructions with narrow focus on the wh-word. Such an analysis of EQ's as narrow focus rather than OQ's captures similarities between EQ's and other instances of focus while accounting for differences between EQ's and OQ's. Moreover, it makes it possible to relate the two most striking properties of EQ's: their unusual obligatory intonation and their extremely restricted discourse distribution.

The intonation contour H*HH% (notation from Pierrehumbert [4]) with the H* accent on the wh-word is obligatory for EQ's such as (1) and quite odd for OQ's.

```
(1)   Fido ate WHAT for dinner?
         H*           H H%
```

EQ's and OQ's are in complementary distribution with respect to context. The questioned element in EQ's must be very salient while for OQ's such salience is infelicitous. In contrast, EQ's and declaratives with the same H*HH% intonation, such as (2), occur in the same contexts and with very similar interpretations.

```
(2)   Fido ate KLEENEX for dinner?
                      H*           H H%
```

If these are both taken to be cases of narrow focus marked by the H*, the minor difference in interpretation follows from the difference in content of *WHAT* and *KLEENEX*. Following Vallduvi (1992), the focus is taken as encoding the information in the utterance. For both EQ's and declaratives of the type in (2), interpretation requires inferring why, when the entire proposition, and in particular the focussed item, are so salient the speaker doesn't 'have' that information. The obvious inferences are that the speaker either didn't hear or didn't believe the prior utterance. The difference in interpretation between the EQ in (1) and the declarative in (2) is that the greater content of *KLEENEX* in (2) makes the "didn't hear" inference less available  [3].

## Extra-propositional Focus and Belief Revision

Keywords: Focus, Belief revision

Sentences such as (1b) with nuclear accent on "believe" pose a problem for current accounts of focus.  In these sentences, "believe" is prosodically marked as though it is the focus but the natural interpretation of the sentences is not that proposed by either semantic (e.g.  Rooth 1985, 1992) or pragmatic (e.g. Vallduvi 1993) accounts of focus.

```
    (1)
a.  SEN. LEAHY:  Did you give a copy to anybody else?
b.  MS. HILL: Other than counsel, I don't believe that I
gave a copy to anyone else.
(Clarence Thomas confirmation hearings, October 11, 1991)
```

The accented "believe" is used to limit the speaker's accountability for the embedded proposition, rather than being part of the core proposition expressed by the utterance.  this "extra-propositionality" is the cause of both semantic and pragmatic accounts' inability to capture the natural interpretation of this type of utterance.  Alternative based semantic accounts fail by incorrectly predicting contrast with propositions in which some other item substitutes for "believe."  The pragmatic accounts fail because the embedded proposition is at least as informative as the accented "believe", contrary to what would be predicted.

As a solution I propose that "extra-propositional focus" explicitly marks its embedded proposition as weaker than the contextually defined default within a system of belief revision  [1],  [8]. The strength of belief associated with a proposition and the speaker's accountability are directly correlated.  In Senate hearings or courtrooms the default for speaker accountability is set at absolute truth.  In less formal situations, the default would be set at a less demanding level and could vary with knowledge of individual speakers' reliability, with speaking style and with the type of interaction.  By using a belief revision framework, the focus-like properties of accented "believe" can be understood as informativeness relative to or contrast with the default belief strength.  In addition, the separation between "extra-propositional focus" and other types of focus can be easily represented by the distinction between the belief strength and associated with propositions and the actual propositions  [2].

## Determiner Ordering

Keywords: Tree Adjoining Grammar, Determiners, English

This joint work is described under the entry for Dania Egedi.

# Beryl Hoffman

Department of Computer and Information Science
hoffman@linc.cis.upenn.edu

## Free Word Order and Combinatory Categorial Grammars

My research concentrates on the syntax and discourse use of word order variation in Turkish. Like Finnish, German, Hindi, Japanese, and Korean, Turkish has considerably freer word order than English. Word order in Turkish is so free that in fact the simple transitive sentence "Chris saw Mary" can be translated to Turkish in six different word orders (i.e. all the permutations of the three word sentence). Word order variation in Turkish and other free word order languages is used to convey distinctions in meaning that are generally not captured in the semantic representations that have been developed for English, although these pragmatic distinctions are also present in somewhat less obvious ways in English. My research intentions are to gain a better understanding of word order variation and to provide better representations of sentence meaning that include the contributions of word order; this is of considerable importance for practical applications in machine translation, machine assisted translation, and computer assisted language training.

To capture the syntactic features of a free word order language, I present an adaptation of Combinatory Categorial Grammars (CCGs) [4] [5] called {}-CCGs (set CCGs) in [3]. {}-CCG is a categorial grammar where a verb's subcategorization requirements are relaxed so that it requires a set of arguments without specifying their linear order. In my dissertation work, I investigate the formal aspects of {}-CCG, namely, its weak generative capacity and its relation to other formalisms such as Linear Indexed Grammar variants.

{}-CCG can be used to parse and generate all word order variations in Turkish sentences; however, it does not capture the more interesting questions about word order variation: namely, why speakers choose a certain word order in a certain context and what additional meaning these different word orders provide to the hearer. To capture these, I integrate a level of information structure, representing pragmatic functions such as topic and focus, with {}-CCGs. I adopt a simple compositional interface in which every {}-CCG category encoding syntactic and semantic properties is associated with a pragmatic category (a component of the information structure); two {}-CCG constituents are allowed to combine to form a larger constituent only if their pragmatic counterparts can also combine. This approach allows certain pragmatic distinctions to influence the syntactic construction of the sentence in a compositional way.

To demonstrate this approach, I present work in progress in [1] on a generation system for a simple database query task which produces Turkish sentences with word orders appropriate to the context. My implementation concentrates on the realization component of generation, using a head-driven bottom-up generation algorithm I have modified for {}-CCGs. In future research, I would like to extend this same approach to generate certain stylistic constructions in English such as topicalization, it-clefts, and right dislocation.

# Paul S. Jacobs

Visiting Associate Professor, Department of Computer and Information Science
pjacobs@unagi.cis.upenn.edu

## Corpus-based Text Interpretation

Keywords: Semantic Interpretation, Lexical Disambiguation, Text Interpretation, Multilingual Text
Processing, Corpus-based Processing

I am at Penn on a one-year faculty appointment, with three basic objectives: (1) through research and course instruction, to cultivate and support corpus-based language work, (2) to participate actively in the government and academic community, particularly in developing an infrastructure for natural language evaluation, and (3) to co-author a new textbook on natural language processing, emphasizing modern themes such as lexicon development, combining knowledge-based and statistical processing, scale-up issues, and current practice.

Word senses are a key source of information for text interpretation. My research has shown that the discrimination of word senses with respect to conventional lexicons is not as helpful as it should be, because real tasks such as data extraction and machine translation demand lexical discrimination along different dimensions. The richest source of information about words and word senses is corpus analysis, thus I have set out to show how word sense knowledge can be acquired from corpus data. This includes, for example, relating words and phrases to their specialized interpretation (e.g. is a *television* something that you watch or an appliance you buy in a store), identifying synonyms and related words in context (e.g. *television* relates to *broadcasting* but also to *set*), and extending such knowledge across languages (The specific senses of *develop* in Japanese have much of the same ambiguity as in English). Some of the results are reported in [2].

During my stay at Penn, I have continued my responsibilities as Principal Investigator of the GE-CMU TIPSTER project [1], which has been the testbed for much of this work. In addition to the TIPSTER program objectives of advancing the art in coverage and accuracy in text interpretation, the program has emphasized evaluations, such as the Message Understanding Conferences (MUC), which are now moving toward including word senses, predicate-argument structure, and reference resolution as well as task-driven evaluation.

The natural language processing textbook, co-authored with Kathleen McKeown (at Columbia University), is due to be published by PWS-Kent in 1995.

# Aravind K. Joshi

Professor, Department of Computer and Information Science
joshi@linc.cis.upenn.edu

## Relationship of Hierarchical Structure and Context-Sensitivity

Keywords: Grammar formalisms, Hierarchical structure, Context-sensitivity

Sentences are not just strings of words (or are they?), they have some (hierarchical) structure. This much is accepted by all grammar formalisms. But how much structure is needed? The more the sentences are like strings the less the need for structure.

A certain amount of structure is necessary simply because a clause may embed another clause, or one clause may attach to another clause or parts of it. Leaving this need of structure aside, the question then is how much structure should a (minimal) clause have? Grammar formalisms can differ significantly on this issue. Minimal clauses can be just strings, or words linked by dependencies (dependency trees), or with rich phrase structure trees, or with flat (one level) phrase structure trees (almost strings) and so on. How much hierarchical structure is needed for a minimal clause is still an open question that is being debated heatedly.

How are clauses put together? Are these operations more like string manipulations (concatenation, insertion, or wrapping, for example) or are they more like tree transformations (generalized transformations of the early transformational grammars, for example)? Curiously, the early transformational grammars, although clearly using tree transformations, actually formulated the transformations as pseudo string-like operations! More recent non-transformational grammars differ significantly with respect to their use of string rewriting or tree rewriting operations.

Grammar formalisms differ with respect to their stringiness or treeness. Also during their evolution, they have gone back and forth between string-like and tree-like representations, often combining them in different ways. These swings are a reflection of the complex interplay between aspects of language structure such as constituency, dependency, dominance, locality of predicates and their arguments, adjacency, order, and discontinuity. These issues are being investigated with respect to a range of recent grammar formalisms.

## Unfolded Types as Building Blocks for Categorial Grammars

Keywords: Lexicalized Grammars, TAG, Categorial Grammar

A lexicalized grammar consists of elementary structures anchored on lexical items and general rules for composing these elementary structures. Context-free grammars (CFGs), in general, are not lexicalized and cannot be lexicalized with substitution alone. Substitution and adjoining can lexicalize CFGs and the resulting system is the same as Lexicalized TAGs (LTAG). In this sense, TAGs arise naturally in the process of lexicalizing CFGs.

The main goal of this work is to define a TAG-like system *entirely* within the framework of categorial systems. In this way, we will be able to combine the key idea in categorial systems, in particular, the tight interface between syntax and semantics, and the key idea in TAGs, namely the extended domain of locality and factoring recursion from the domain of dependencies.

First, I will describe some background and then state some of the key ideas of this new work. There are many interesting relationships between TAGs and Categorial Grammars (CG). For example, Weir (1987) has shown that TAGs are equivalent (with respect to the weak generative capacity) to Combinatory Categorial Grammars (CCG) of Steedman, under certain conditions. Vijay-Shanker and Weir (1990) developed a common parsing architecture for TAGs, CCGs, and Linear Indexed Grammars (LIG), indirectly based on the equivalence of these systems.

We can describe more fine-grained relationships between TAGs, more specifically Lexicalized TAGs (LTAGs) and CCGs, based on the observation that both LTAGs and CCGs are lexicalized grammars. LTAGs (with substitution and adjoining) are similar to CCGs in the sense that, for each lexical item, the elementary tree(s) in an LTAG, which is (are) anchored on that item can be regarded as the *structured* category (categories) associated with that item. One of the elementary trees associated with *likes* is (represented in a labeled bracketed form, anchored on V) S[ NP VP[ V NP]]. (This is just one of the trees associated with *likes*). In a CCG, the syntactic type associated with *likes* is (S\ NP)/NP. The CCG representation and the LTAG tree for *likes* both encode the information that *likes* has two NP arguments. However, the LTAG tree also encodes the structural positions for the two arguments. It also encodes a specific CCG derivation. Further, the LTAG tree makes a commitment to certain constituencies. In a CCG, there is a strict correspondence between types and constituents, i.e., each type is a constituent and each constituent is a type. In fact, this property is exploited by CCG in its novel account of coordination. This is not the case for LTAGs. For each lexical string built by the operations of substitution and adjoining, there is obviously a functional type that can be read from the elementary or derived tree. Thus, for *likes,* it is NP x NP → S, for *likes peanuts*, it is NP → S, and for, *John likes*, it is NP → S. The first two strings are constituents but the last one is not. Hence, in LTAGs the type-constituency correspondence is not strict. It is partial. In LTAGs, constituencies are defined at the level of the elementary trees, no other constituencies are introduced during the derivation. However, every string has a functional type associated with it.

Based on the above considerations Joshi and Schabes (1991) showed how a CCG-like account of coordination can be given in LTAGs. The coordination schemas are defined over the *structured* categories. In particular, Joshi and Schabes showed how an account parallel to Steedman's treatment of coordination and gapping can be given in LTAGs. There are some interesting differences in these two treatments but they are essentially parallel. Although this work clearly shows some close relationships between LTAGs and CCGs, they do not allow a direct comparison of LTAGs and CCGs.

The key idea in constructing a TAG-like system entirely within the categorial framework is to assign elementary *partial* proofs (proof trees) of certain kinds as types to lexical items rather than the types associated in a categorial grammar. These partial proofs will include *assumptions* (assumption nodes) which must be *fulfilled* by *linking* the conclusion nodes of partial proofs to assumption nodes. Roughly speaking, these partial proofs are obtained by unfolding the types associated with the lexical items. This allows us to associate an extended domain of locality to the structure associated with a lexical item, analogous to the trees of LTAGs.

Partial proofs are *composed* to obtain proofs for strings of lexical items. We need to go further however. Treating a node of a proof tree as a pair of conclusion and assumption nodes, a proof tree can be *stretched*. Then an appropriate proof tree can be *inserted* by linking conclusion nodes to assumption nodes.

During unfolding the syntactic type associated with a lexical item by a categorial grammar, we will also allow *interpolation*. That is, during unfolding, we can interpolate a proof. Interpolation is like stretching except that unlike stretching, the interpolated proof has to uninterpolated by linking it to a non-null proof tree.

CCGs have no fixed constituencies, LTAGs have fixed constituencies defined at the level of elementary trees. To capture this property, we need to consider assumptions that are really *traces*. These trace-assumption nodes are *discharged* internally (locally) in the elementary proof trees. Discharging these trace assumptions is exactly like discharging assumptions in a natural deduction system. Only trace assumptions are discharged in this way. The assumptions we talked about earlier are not discharged. They have to be fulfilled by linking them to conclusion nodes of other partial proof trees. The discharge of trace assumptions locally within an elementary tree not only allows us to define fixed constituencies but also to capture long-distance dependencies in a *local* manner analogous to their treatment in LTAGs.

In summary, the elementary proof trees associated with a lexical item are constructed by unfolding the syntactic type up to atomic types. While unfolding, we can optionally stop if the conclusion is the same as one of the arguments, i.e., assumptions. If trace assumptions are introduced then they have to be locally discharged, and finally, while unfolding a proof tree can be interpolated. Proof trees are combined with proof trees by linking, by stretching and linking, and by uninterpolating by linking. This system appears to be adequate to describe the range of phenomena covered by the LTAG systems and the corresponding weakly equivalent categorial systems such as the Combinatory Categorial Grammars of Steedman. If we are successful in constructing a system as described above then there is possibility of extending the parsing algorithms for LTAGs to this system, thus achieving polynomial parsability.

Another way of viewing this work is as follows. Starting with CFGs, by extending the domain of locality, we arrive at LTAGs. Starting with Categorial Grammars (the so-called Bar-Hillel–Ajdukiewicz grammars, BA), by extending the domain of locality as described above, we arrive at the system described above.

# Jonathan M. Kaye

Department of Computer and Information Science
kaye@linc.cis.upenn.edu

## Trauma Modeling of Anatomy and Physiology

Keywords: Qualitative simulation, Spatial reasoning, Medical informatics

For the past three years, I have been working with Dr. Bonnie Webber and John Clarke, M.D. on the TraumAID project [10]. For my initial exposure, I ported TraumAID from the Symbolics machine to the X-Windows and Macintosh environment, in addition to reworking it for easier transport to any other platform. Understanding TraumAID better, I came to recognize deficiencies that were common to other programs in Medical Informatics. My interest in causal modeling as a means for sound explanation led me to my dissertation topic: reasoning about the effects of spatial disruptions from trauma in human anatomy with knowledge about physiological processes.

I see two principal directions in which medical programs have emerged: analysis of space (anatomy) and of function (physiology and pathophysiology). While many domains involve both, typically a program only focuses on one.

It is not hard to see that these directions ultimately will meet. At the same time, there have been few attempts in medicine to integrate space and function meaningfully [3], in spite of its importance [1]. Researchers in Medical Informatics are recognizing the value of designing knowledge for programs to use over a range of applications, rather than on an application-specific basis [4], [9], [7]. I believe that anatomic knowledge has a great potential for reuse because of its central role in medicine.

I am developing a system to integrate spatial (anatomic) and functional (physiological) knowledge about the human body, centering on how structural change due to trauma (initially penetrating injury such as caused by guns and knives) affects physiology. Using Jack [2], I am building a graphical interface to my system to provide an intuitive user interface, an illustration of the system's knowledge about the situation, and a tool to assist medical professionals in visualizing the internal extent of injury they cannot see directly.

Based on clinical findings and tests, the system will try to arrive at hypotheses about the potential for injury due to spatial and functional constraints. It will construct qualitative models of simplified, acute physiological systems and enable users to simulate how spatial changes affect physiological parameters. Currently, I am using QobiSim [8] as the simulation engine, based on the QSIM [5] paradigm, and anticipate using Metaxas' deformable models [6] to model anatomic parts. As the first step, we are constructing models to represent the breathing mechanism and stable acute cardiovascular system.

I expect that implementing such a system will serve as a starting point for linking medical imaging and functional analysis. I envision its direct impact as reinforcing the fundamental interaction of anatomy and physiology, to give the computer a solid framework for representing medical knowledge. This could aid in learning about their relationship, solidifying the medical professional's mental image of the situation, and reusing anatomic knowledge.

# Albert Kim

Department of Computer and Information Science
alkim@unagi.cis.upenn.edu

## Interactive Processing using Graded Unification

I am interested in "interactive" models of human sentence processing. Specifically, I am attempting to model the employment of lexical semantic information in the resolution of ambiguities like the following:

1. The man recognized by the spy took off down the street.

2. The van recognized by the spy took off down the street.

Recent psycholinguistic research such as Trueswell *et al.* [7] demonstrates rapid employment of thematic role information in sentence processing using materials like (1) and (2). Eye-tracking shows that subjects resolve the ambiguity rapidly (before reading the *by*-phrase) in (2) but not in (1). The conclusion Trueswell *et al.* draw is that subjects use knowledge about thematic roles to guide syntactic decisions. Since *van*, which is inanimate, makes a good Theme but a poor Agent for *recognized*, the past participial analysis in (2) is reinforced and the main clause (past tense) suppressed. Being animate, *man* performs either thematic role well, allowing the main clause reading to remain plausible until the disambiguating *by*-phrase is encountered. At this point, readers of (1) display confusion.

Modelling such performance requires an outlook which is at least slightly more flexible than the classical linguistic focus on well-formedness. Trueswell *et al.* [7] point out that the processing effects observed in subjects are graded, and appear to correlate to the degree of "thematic fit" between verbs and their complements. In general, the various pieces of non-syntactic information available to the processor may conspire to support the same hypothesis or may support competing hypothesis; the "correct" hypothesis will not be supported by all the evidence, but by the most or the most credible.

In order to accommodate a more dynamic flow of information in language processing, I have developed an extension to classical unification (see [5]), called *graded unification* [3]. Graded unification is capable of combining two feature structures which are incompatible under classical unification, and returns a strength which reflects the compatibility of the combined structures. This new operator is similar in spirit to the operators of fuzzy logic (see [2]). I have also developed a parser based upon graded unification. This parser is capable of employing graded constraints on combinations as well as frequency information to guide its decisions.

The lexical nature of the phenomena which interest me suggests working within a highly lexicalized grammatical framework such as CCG or HPSG. I have recently become attracted to CCG, and have joined an effort to develop a wide coverage CCG grammar and parser [1]. The project has bootstrapped a CCG grammar and statistics about the use of its categories from analogous information collected for the Tree Adjoining Grammar (TAG) of the XTAG project. This bootstrapped grammar is intended to drive a graded-unification-based CCG parser which is currently under development.

The next and most important stage of my work involves enriching a grammar with semantic information to guide the graded-unification-based parser. I am currently trying to identify a set of salient semantic "features" which are important in determining thematic fit. One idea, which I am investigating in work with John Trueswell of the Psychology Department, is to use the the *semantic differential technique* [6] to draw a semantic space occupied by concepts (in this case, nominal concepts). Our tentative plan is to focus on a select group of interesting verbs, such as *examined*, occurring in a richly annotated corpus. We intend to automatically acquire the semantic constraints these verbs impose upon their thematic role fillers (their complements). The training data will be developed from the annotated corpus, which will help identify the thematic role of each complement, and from the application of the semantic differential technique, which will semantically characterize each complement. We are aided in this endeavor by a recently enriched annotation scheme adopted by the Penn Treebank project [4].

# Nobo Komagata

Department of Computer and Information Science
komagata@linc.cis.upenn.edu

## Formal Properties of CCG

Keywords: Categorial Grammar, Formal Properties, $\lambda$-Prolog

### Formal Properties of CCG with Lexical Type Raising

Previous results show that:

1. Combinatory Categorial Grammar (CCG) with bounded functional composition is equivalent to Linear Index Grammar (LIG) [4],

2. if this boundedness condition is removed, it is more powerful [4],

3. the addition of conjunction to bare CCG results in more powerful languages [4], and

4. the use of variable (e.g., V/a/(V/b)) can generate non-Linear Index languages [1].

I am trying to show that there is a variant of CCG with lexical type raising [3] which is still equivalent to LIG.

### English and Japanese Lexicon

A more linguistic side of my study is about computational representation of the categorial lexicon with data drawn from English and Japanese. Since these languages have contrasting features, it is interesting to see what role linguistic universals play in the representation.

### $\lambda$-Prolog Implementation

I am exploring the possibility of using $\lambda$-Prolog as the language of implementation. The language provides a straightforward representation of the higher-order semantics of type raising. Its strong type system is another attractive feature to better organize the objects and the operations.

# Seth Kulick

Department of Computer and Information Science
skulick@linc.cis.upenn.edu

## Resource-Boundedness in Formalisms for Natural Language Processing

Keywords: Linear Logic, TAG, Categorial Grammar

I am currently investigating the mathematical and computational aspects of the role of "resource-boundedness" in formalisms used for natural language processing. The two formalisms that are the focus of the research are Lexicalized Tree Adjoining Grammar (LTAG) and Combinatory Categorial Grammar (CCG). LTAG and CCG are both lexicalized formalisms, meaning that each entry in the lexicon is not just a single word, but rather a structure requiring certain "resources" from other lexical entries in order to be completed. Thus, from the perspective of resource-boundedness, both formalisms have some fundamental similarities. There is also a similarity to linear logic, a branch of logic in which a premise is not available forever once it is introduced, but rather is treated as a resource to be used in a finite manner. Such a resource-oriented viewpoint has been shown to have worthwhile linguistic applications [1]. Just as resource-boundedness is explicitly built into the framework of linear logic, these grammar formalisms may potentially be altered to make the resource aspects more explicit. This would allow a unification of what had previously appeared to be quite different aspects of these grammars, thus expanding on work previously done in this direction [2]. Ideally, this will result in a grammar formalism that combines the advantages of both LTAG and CCG, in terms of linguistic coverage and processing complexity.

It is possible that such consideration of resource-boundedness may be relevant not only to syntax but also to discourse analysis. The problem of introducing entities in discourse and then referring to them by definite description as pronouns can also be considered as constraints on resource-boundedness. An entity introduced in discourse can be referred to within the boundaries of a certain discourse structure but once outside this structure the entity needs to be introduced again in order to be referred to by a pronoun. These considerations are similar to the limited use of a premise and the reintroduction of the same premise for future use. I plan to cast the use of these discourse constraints into the framework of resource-bounded logics and investigate the characteristics of discourse structure in this framework. Hopefully, these investigations will suggest some new discourse processing strategies.

# Sadao Kurohashi

Visitor, Institute for Research in Cognitive Science
sadaok@linc.cis.upenn.edu

## Toward Real Natural Language Understanding

Keywords: Coordinate structure, Case structure, Discourse

I am interested in practical ways of processing natural language text. With a colleague at Kyoto University, I have developed several novel methods and algorithms for analyzing Japanese sentences [3]. These methods are:

- Japanese Morphological Analyzer, JUMAN.

- Coordinate Structure Analysis
  In a coordinate structure, there exists a certain, but sometimes subtle, similarity between conjuncts. By measuring such subtle similarities between two arbitrary series of words, the proper boundaries of coordinate structures can be detected [1].

- Case Structure Analysis
  Structural ambiguity and word sense ambiguity are intrinsic obstacles to case structure analysis. I have developed a method of resolving these ambiguities simultaneously by matching an input sentence with the typical example sentences given in a case frame dictionary, resulting in the case structure representation of the sentence [2].

- Discourse Structure Analysis
  In the case of scientific and technical texts, considerable part of the discourse structures can be estimated automatically by incorporating the three types of clue information: clue expressions, word/phrase chains, and similarity between sentences [4].

At present I am evaluating these methods through experiments on large amounts of texts. Analyzing the results will suggest other interesting targets for real natural language understanding.

In addition, I am also trying to apply the coordinate structure analysis method to English. I believe that it will work well for English coordination with only minor modifications.

**Libby Levison**

Department of Computer and Information Science
libby@linc.cis.upenn.edu

## Object Specific Reasoning

General-purpose AI planners, in decomposing task plans into the steps which comprise the plan, have tended to produce steps which lack manipulation details and are thus too general to be executed. Robotic applications, meanwhile, have tended towards building domain-specific manipulation procedures which require that that each motion be completely specified by the operator and which do not readily generalize. While human agents can interpret a command such as "pickup glass" – supplying information such as the hand to use and how high to pickup the glass, regardless of the configuration of the environment – current synthetic agents cannot.

This suggests the need for an intermediate reasoning system which can determine the missing details. My research is to build such a system, the Object Specific Reasoner (OSR), which tailors the steps of high-level plans to the specifics of the agent and objects. As plans are elaborated, the OSR generates a sequence of action directives which are ultimately sent to the agent controller. In the process of generating the action directives and grounding the task in the world, the OSR also gives a measure of the feasibility of each action the agent is asked to perform.

In addition to the recognition of the need of such an intermediate reasoner for agent-object interaction, the major contributions of this work lie in the specific architecture. In refining high-level action descriptions, the OSR accepts succinct action requests, makes use of minimal symbolic knowledge, and relies on 'simulated' sensors to gather specific knowledge.

### SodaJack

A prototype of the OSR is part of the *SodaJack* system. Built in conjunction with Chris Geib's high-level planner ITPLANS, Mike Moore's search planner, and Tripp Becket's Behavioral Simulator, the SODAJACK system depicts an animated agent who works the fountain of an old fashioned soda shoppe. The waiter takes 'menu' orders such as (**serve soda**); Geib's ITPLANS selects a plan to accomplish such goals, and decomposes the plan into steps. Moore's search planner identifies and locates objects referred to in the plans; the OSR then takes each plan step and expands it into a set of fully parameterized action directives with which Becket's animation simulator is invoked.

## D. R. Mani

Department of Computer and Information Science
mani@linc.cis.upenn.edu

## The Design and Implementation of Massively Parallel Knowledge Representation and Reasoning Systems: A Connectionist Approach

Systems that model human cognition must use massive parallelism in order to react in real-time. Connectionist models, with their inherent parallelism, seem to be promising architectures for modeling cognition. In exploring such architectures, an understanding of real-time reasoning over a large body of knowledge would offer significant insight into the cognitive as well as practical aspects of knowledge representation and reasoning.

This research investigates mapping structured connectionist models onto existing general purpose massively parallel architectures with the objective of developing and implementing practical, real-time connectionist knowledge base systems [1]. SHRUTI, a connectionist knowledge representation and reasoning system which attempts to model reflexive reasoning [2], [3], will serve as our representative connectionist model. Efficient simulation systems for SHRUTI have been developed on the Connection Machine CM-2—an SIMD architecture—and on the Connection Machine CM-5—an MIMD architecture. The resulting systems are being evaluated and tested using large, random knowledge bases with up to half a million rules and facts.

Though SIMD implementations on the CM-2 are reasonably fast—requiring a few seconds to tens of seconds for answering simple queries—experiments indicate that message-passing MIMD systems (on the CM-5) are vastly superior to SIMD systems and offer ten- to hundred-fold speedups. This greatly improved performance, in conjunction with the fact that the CM-2 is fast becoming obsolete, has resulted in most of the research effort being expended to improve and extend SHRUTI-CM5—the asynchronous MIMD message passing simulation system on the CM-5.

The following are some of the salient features of SHRUTI-CM5 [4]:

**Granularity** In order to provide appropriate granularity, the knowledge base is partitioned at the coarse-grained *knowledge-level* where knowledge elements like predicates, concepts, facts, rules and *is-a* relations form the primitives.

**Mapping** These knowledge-level primitives are mapped onto processors using one of a variety of processor assignment schemes. Random processor assignment provides reasonably good static and dynamic load balancing.

**Knowledge encoding** SHRUTI-CM5 reads and encodes pre-processed knowledge bases in parallel. Each processor independently and asynchronously encodes the fragment of the knowledge structure assigned to it. As the knowledge base is read, internal data structures are constructed to represent the knowledge.

**Simulation** Queries can be posed after the knowledge base has been encoded. During an episode of reasoning, processors asynchronously maintain and update activation frontiers resulting in spreading activation. Computation and communication are overlapped by using non-blocking *active messages* for all inter-processor communication.

**Performance** SHRUTI-CM5 has been tested using knowledge bases with up to half a million rules and facts. With randomly generated, structured knowledge bases, the system running on a 32 node CM-5 can process a variety of queries in well under a second. The queries used had inference depths ranging from 0–10.

This work provides some new insights into the simulation of structured connectionist networks on massively parallel machines and is a step toward developing large yet efficient knowledge representation and reasoning systems. Using the resulting system as a simulation tool, psychologically significant aspects of reflexive reasoning will also be explored. We hope this will further our understanding of the nature of reflexive reasoning and help us evaluate its practical and cognitive significance.

Research advisor for the work described here is Dr. Lokendra Shastri.

# Mitch Marcus

Professor, Department of Computer and Information Science
mitch@linc.cis.upenn.edu

## Deducing Linguistic Structure from Large Corpora

### Automatic Acquisition of Linguistic Structure

Within the past several years, a widening circle of researchers have begun to investigate a new set of techniques for the use of trainable systems in natural language processing. The early successes of these new techniques, coupled with other advances, have allowed the emergence of a new generation of systems that both extract information from and summarize pre-existing text from real-world domains.

A group of us at Penn have initiated a research program to see how far the paradigm of trainable systems can take us towards the fully automatic syntactic analysis of unconstrained text and towards the automatic acquisition of grammatical structure from both annotated and unannotated text corpora. This research is investigating both statistical and symbolic learning methods using both supervised and unsupervised approaches.

Fundamental to our project is an attempt to unite different linguistic traditions often viewed as mutually exclusive. Thus, this work aims to combine the research program of generative grammar, as set forth originally by Noam Chomsky, and the research paradigm of distributional analysis, as developed by the American structural linguists resulting in the mathematical and computational work of Zellig Harris. For an overview of this point of view, see [3]. Similarly, our approach to language learning rests on the premiss that, in addition to exploiting a core of fundamental linguistic properties shared by every language, learners must also employ the technique of distributional analysis to discover a very wide range of potentially idiosyncratic language-particular linguistic phenomena.

### Stochastic Parsing

In an experiment two years ago, we investigated how distributional facts can be used to choose between the multiple grammatically acceptable analyses of a single sentence. The resulting parser, Pearl, [2] differs from previous attempts at stochastic parsers in that it uses a richer form of conditional probabilities based on context to predict likelihood. Tested on a naturally-occurring corpus of sentences requesting directions to vary locations within a city (the MIT Voyager corpus), the parser correctly determined the correct parse (i.e. gave the best parse first) on 37 of 40 sentences. We are now beginning a collaboration with the Continuous Speech Recognition Group at IBM's Thomas Watson Laboratory to develop a new generation of stochastic parsers, based on decision tree technology utilizing a rich set of linguistic predicates, and trained on output from both the Penn Treebank (see below) and the Lancaster Treebank. (A first version of such a parser [1] developed at IBM, with Magerman's participation, can be viewed as an extension of Pearl.)

**The Penn Treebank Project**

We have been working on the construction of the Penn Treebank, a data base of written and transcribed spoken American English annotated with detailed grammatical structure. This data base, although now only in preliminary form, is serving as a national resource, providing training material for a wide variety of approaches to automatic language acquisition, a reference standard for the rigorous evaluation of some components of natural language understanding systems, and a research tool for the investigation of the grammar and prosodic structure of naturally spoken English.

The Penn Treebank project has just completed its first, three-year phase. During this period, 4.5 million words of text were tagged for part-of-speech, with about two-thirds of this material also annotated with a skeletal syntactic bracketing. All of this material, now available in preliminary form on CD-ROM through the Linguistic Data Consortium (LDC), has been hand-corrected, after processing by automatic tools. The largest component of the corpus consists of materials from the Dow-Jones News Service; over 1.6 million words of this material has been hand parsed, with an additional 1 million words tagged for part of speech. Also included is a skeletally parsed version of the Brown corpus, the classic million word balanced corpus of American English. This corpus has also been hand-retagged using the Penn Treebank tag set. Smaller tagged and parsed subcorpora include 100K words of materials from past ARPA Message Understanding Conference (MUC) and 10K words of sentences from the ARPA-sponsored Air Travel Information System (ATIS) spoken-language system project.

The error rate of the part of speech tagged materials (done several years ago) is estimated at approximately 3%. About 300,000 words of text have been corrected twice (each by a different annotator), and the corrected files were then carefully adjudicated, with a resulting estimated error rate of well under 1%. All the skeletally parsed materials have been corrected once, except for the Brown materials, which were instead quickly proofread an additional time for gross parsing errors.

Earlier material, released through the ACL/Data Collection Initiative, has been used for purposes ranging from serving as a gold-standard for parser testing, to serving as a basis for the induction of stochastic grammars (including work by groups at IBM, and a collaboration between Penn, AT&T Bell Labs and Harvard University), to serving as a basis for quick lexicon induction for the MUC task (in unpublished work at BBN.)

The Penn Treebank Project, now in its second phase, is working towards providing a 3 million word bank of predicate-argument structures. This is being done by first producing a corpus annotated with an appropriately rich syntactic structure, and then automatically extracting predicate-argument structure, at a level of detail which distinguishes logical subjects and objects, as well as distinguishing arguments from adjuncts (for clear cases). This syntactic corpus will be annotated by automatically transforming the current Penn Treebank into a representational structure which approaches that of the intended target, and then completing the conversion by hand. The preliminary version of the corpus is being substantially cleaned up at the same time. The second release of the Penn Treebank should be available through the LDC in late 1993. For more information, see [4].

# I. Dan Melamed

Department of Computer and Information Science
melamed@linc.cis.upenn.edu

## Corpus-Induced Semantic Maps

My long-term research goal is to develop generic methods of finding mappings between different representations of semantics. The semantics can be those of natural languages, computer languages or a combination. I am using statistical methods to derive these mappings from large online corpora, to ensure that the results degrade gracefully with broader coverage.

Since there are no large bilingual corpora of semantic representations, I am extracting the next best thing from the online Canadian Hansard proceedings. Using statistical models and various online tools and resources, I am experimenting with various algorithms for machine translation. These algorithms are neither purely statistical, nor purely representational, but hybrids of the two complementary methodologies.

Such algorithms can be applied equally well to non-natural languages. I apply them to the NLP component of the ARPA sponsored ATIS task. ATIS is a hypothetical voice-operated Air Traffic Information System. The output of the system is SQL code for the Online Airlines Guide (OAG) database. Although purely representational methods of translating from English to SQL have been relatively successful, they are not easily portable to non-database arenas. This task is a natural training ground for algorithms that translate from natural language to computer code, because "bilingual" English - SQL corpora are available online. Furthermore, the Penn Treebank contains part-of-speech and parsing information for the English half of some of these corpora, which provides even greater opportunity for experimentation with different techniques.

Many problems need to be solved before such semantic maps can be charted. My current strategy is to write multi-stage bottom-up algorithms. In a probabilistic context, this means that a map is constructed by a series of successive approximations. The first approximation is generated from the simplest statistical model, which takes into account only very broad outlines of the semantic representations involved. Later "stages" would improve on this first guess, using larger models, and considering finer details of the semantics. The details of each part of the algorithm still need to be worked out. Then I shall experiment with ways to merge the successive approximations to produce an optimal maps.

Hybrid semantic mapping techniques need not be limited to natural or computer languages. In theory, they can also be applied to more abstract semantics, or semantics yet to be defined, such as the semantics of perception, the semantics of action or the semantics of logic. This can lead to new ideas in robotics, computer graphics, programming languages, and human-computer interaction.

# Michael B. Moore

Department of Computer and Information Science
mmoore@linc.cis.upenn.edu

## Plans for search behavior

Keywords: Planning, Search

People often do not know where things are and have to look for them. My research presents a formal model suitable for reasoning about how to find things and acting to find them, which I will call "search behavior". Since not knowing the location of something can prevent an agent from reaching its desired goal, the ability to plan and conduct a search will be argued to increase the variety of situations in which an agent can succeed at its chosen task.

Searching for things is a natural problem that arises when the blocks world assumptions (which have been the problem setting for most planning research) are modified by providing the agent only *partial* knowledge of its environment. Since the agent does not know the total world state, actions may *appear* to have nondeterministic effects. The significant aspects of the search problem which differ from previously studied planning problems are the acquisition of information and iteration of similar actions while exploring a search space.

In Moore93 [2] I outlined three alternatives to generating search behavior. These three approaches produce the same sequence of actions, but differ to the degree that they represent a complete plan for conducting a search. The simplest approach uses a sequential planner as a subroutine in a heuristic search algorithm. Each sequential plan produced is incomplete, since it only represents a sequence of actions and leaves implicit the conditional branching inherent in the search algorithm. The other two approaches explicitly represent conditional branches and iteration, respectively.

The last of these approaches has been implemented as part of the AnimNL project's SodaJack system (described in GLM94 [1]). All the search planners are also being implemented in a testbed environment derived from Polalck's [3] TileWorld testbed for rational agents. These two testbed environments: the AnimNL simulator and TileWorld are being used to gather comparative performance data.

# Charles L. Ortiz, Jr.

Department of Computer and Information Science
clortiz@linc.cis.upenn.edu

## The preference problem for counterfactuals

Keywords: Counterfactual reasoning, reasoning about action, belief revision.

The *preference problem* for counterfactuals [5] is the problem of preferentially ordering possible worlds in order to accommodate some counterfactual supposition. Contemporary treatments of counterfactuals in AI have been limited to preferential orderings based simply on, for example, cardinality of sets of differences between competing worlds (see, for example, [2], [6], [1]). However, such naive preferences can enjoy only limited success as soon as one's ontology is expanded to include events and time and as soon as the possibility of ramifications to one's actions are admitted. I claim that instead possible worlds can be ordered along a number of ontological dimensions: actions, facts, defaults, and time; and further that knowledge of the performing agent's mental state — in the form of the agent's intentions as well as what I term a situation-evoked *action context* — further constrain the set of closest possible worlds.

Counterfactual reasoning is regarded as central to commonsense reasoning in AI: for example, to planning and diagnosis. In [4] I claim that counterfactual reasoning should as well play a central role in our commonsense conceptualization of action: many act-types are best analyzed in terms of counterfactuals. In [3] and [4] I present a formal analysis based on counterfactuals of notions of prevention, enabling, and maintaining, as well as an analysis of Goldman's notion of generation. This analysis raises a number of ontological questions regarding distinctions between an agent's basic movements and his high-level actions and their role in the evaluation of counterfactuals. It has also motivated a proposed program — involving generating responses to *what-if* and *how* queries — to test and further refine my theoretical claims.

# Martha Palmer

Adjunct Professor, Department of Computer and Information Science
mpalmer@linc.cis.upenn.edu

## Issues in Lexical Semantics

I spent many years at Unisys helping to develop the PUNDIT text analysis system [12]. One of the biggest bottlenecks in porting Pundit to new domains was the difficulty of entering new lexical semantic information. The reason for this was partly theoretical, and partly a software development issue. As the system had grown, and different components had been added, the entering of semantic information had not been coordinated at one site. Each component—the parser, the lexicon, the semantic analysis routine, and the temporal analysis routine—required different kinds of semantic information about the same lexeme, each of which had to be entered separately. On the surface, this could have been solved by a global lexical entry procedure that would automatically perform the tedious data entry necessary, putting everything in the right place. In fact, some types of information, such as semantic class co-occurrence patterns and verb-argument selection restrictions, or the verb prepositions doubly listed in the lexicon and in the linking rules, were completely redundant, and could have been collapsed. But in other cases, the entry problems were symptomatic of a more serious flaw in the system. We did not have a theory of verb representation that adequately accommodated all the necessary information. The predicate argument structure of a verb was represented in the form of inference rules that could not be modified with features or properties such as temporal information. We also had no way of making generalizations about optional situation adjuncts, such as locatives. These are arguments that are associated with classes of verbs, such as Source and Goal for MOTION verbs [7]. We had no way of representing argument structure for a class of verbs rather than an individual verb.

In addition to the verb representation making it difficult to generalize about verb classes, the semantic analysis routine that used them was inflexible with regard to satisfaction of semantic constraints. Constraints either succeeded or failed. The verb representations were not "robust" in the sense that they could not partially match a syntactic parse or choose between competing partial matches. They could not recognize minor extensions of verb meaning. *Broken starting air compressors* were represented as "inoperative" or "nonfunctioning." *Broken wire* required a new sense of BREAK and was represented as "separated into pieces." *Broken insulation*, even though syntactically similar to *broken wires*, required a third sense, with a semantics including a representation of "non-functional and partial separation" [11].

A conceptual taxonomy could store lexical-semantic information more coherently and more flexibly, by capturing syntactic, lexical, selectional, and pragmatic generalizations about necessary and optional verb arguments at either the verb level or the verb class level. This would simplify lexical acquisition, since close conceptual "neighbors" provide the most likely extensions of a verb meaning [14]. They would also provide indications of which directions should first be explored when trying to account for new, previously undocumented usages of a verb which would greatly improve robustness.

My current research is to define such a taxonomy. Recent, extensive studies of English verb classifications such as Levin's [9] and Jackendoff's [7] lay a foundation for further, more specific work on verb classification, and, perhaps more importantly, outline the beginnings of a methodology for pursuing such work. It has been demonstrated that looking at the same verb in different languages, specifically French and Mandarin, can be very helpful in determining the relevant conceptual structure needed for computational applications [13]. Mandarin, in particular, requires a more explicit representation of both the action involved and the resulting state after a BREAK event. *Break* in *John broke the branch* is translated as *da duan—break into line segments*, while in *John broke the glass* the preferred translation is *da sui—break into many small pieces*. In addition, Beth Levin's methodology for using syntactic alternations for distinguishing between verb classes can be extended to the isolation of separate senses of the same verb. The primary sense of BREAK, 'separate into pieces,' can take an optional Instrument argument, as in *John broke the door with a crowbar.* However, a secondary sense, 'become nonfunctional' as in *The computer is broken*, cannot take an Instrument argument, or at least cannot take a Tool as an Instrument. The addition of the Tool as the Instrument forces the selection of the primary sense, as in *John broke the computer with a hammer* [13].

It is anticipated that Levin's methodology will continue to be critical to the development of this verb taxonomy. An important element of the current proposal is to apply what is at this point essentially a linguistic methodology for verb classification to a computational setting. Here the validity of the approach can be tested by incorporating it into a system (including the necessary extensions to overcome the shortcomings of the linguistic system). In addition to Levin's classification, other important sources of data for this taxonomy will be lexical ontologies such as the ones used in large-scale MT systems at ISI [6], CMU [10] and the University of Maryland [2].

I am collaborating on a cross-linguistic verb taxonomy for machine translation [5] with Bonnie Dorr and Patrick St. Dizier, where we are also examining French and Korean. One application of this taxonomy is a Korean-English Machine Translation project using Synchronous TAGS [4] [8] that is being developed here at Penn in conjunction with Dania Egedi and Hyun Seok Park.

See also the note on the STAG Machine Translation Project.

# Hyun S Park

Department of Computer and Information Science
hspark@linc.cis.upenn.edu

## Korean Grammar Using TAG

I am currently investigating various issues related to representing the grammar of Korean for machine translation using TAG. This includes building elementary trees for Korean, tackling the scrambling problem, recovering empty arguments, and comparing between CCG and TAG. I am particularly interested in whether pragmatic, semantic, and processing constraints should be reflected in the Korean TAG syntactic representation.

### Korean Grammar

I am attempting to develop a Korean grammar along the lines of that seen in the old XTAG technical report [1]. It will be entitled *A Lexicalized Tree Adjoining Grammar for Korean*. It will illustrate all of the elementary trees in the Korean grammar, along with an explanation of the Korean grammar itself. There are three structures for Korean: a basic structure, a modified structure, and a conjunctive structure. I'll focus on the modified structure. The Korean causative structure has received many different interpretations from various schools of linguistic theory. It is represented differently from the other biclausal structures in TAG notation, and this will be carefully studied.

### Machine Translation

I have been working with Martha Palmer and Dania Egedi on machine translation using TAG for the last few months. The current system is based on using the Synchronous Tree Adjoining Grammar formalism which is an extension of lexicalized, feature-based Tree Adjoining Grammars (FB-LTAGs). Here, we are investigating how the basic idea of Synchronous TAG can be used for Korean to English and English to Korean translation. We hope to account for Wh structure and relative clause structure for both languages and how the translation of the structures can be mapped. However, we'll not be dealing with structures with scrambling or empty arguments here for the moment, as handling such a structure for both languages means a dramatic changes in TAG notation. Our work will be initially focused therefore on lexical selection and feature transfer.

### Scrambling

I am investigating scrambling in Korean and the problems with the previous TAG representation. Previous work done at Penn using Multi-Component TAGs [4] may prove useful, as the multi-component TAG approach may be appropriate for handling both scrambling and empty arguments in Korean.

### Recovering Empty Arguments

I am currently working on a theory of recovering empty arguments in Korean. Recovering empty arguments is a crucial factor for interpretation as well as parsing. This is especially important in our current work in machine translation which involves army telecommunication discourse messages. I am focusing on semantic features in the initial work; however, as our model is a discourse model, pragmatics will be dealt with in the future. We hope to come up with a formula that combines the TAG syntactic tree notation together with semantic features and pragmatics that handles empty arguments.

**CCG vs TAG**

Another interesting area is comparing Combinatory Categorial Grammar with TAG in handling scrambling for Korean. Both CCG and TAG have been proven to be in the class of Mildly Context Sensitive Languages [2]. However, the current approach to handling scrambling is completely different in CCG, which is basically based on work by Young-Suk Lee and Michael Niv [3]. I'll explore how CCGs can be used to produce the same Korean grammar represented by TAG. Currently, only a small portion of the Korean grammar is implemented in CCG, but I am working on extending it.

# Jong Cheol Park

Department of Computer and Information Science
park@linc.cis.upenn.edu

## Quantification, Argument Structure and CCG

Traditional approaches to quantifier scope ambiguity have utilized such apparatus as quantifying-in, quantifier raising, or a free variable constraint in order to generate all semantic forms that correspond to only *available* readings associated with natural language sentences. In my work to be reported in my dissertation proposal [1], I have shown that quantifier scope has a property such that grammatical scopings of quantifiers in a natural language sentence are *localized* with respect to its function-argument structure, once other semantic ambiguities are factored out. Apparent exceptions to this property are shown to be limited to the *referential* uses of quantified NPs. According to this property, traditional accounts of quantifier scope not only overgenerate available semantic forms: They fail to account for the nature of quantifier scope in natural language.

Function-argument structures are not only useful in making explicit the locality property of quantifier scope. They are also required at some stage of natural language understanding for a truth-conditional evaluation of natural language sentences. They also have sufficient information for a reasonable specification of binding relations, which in traditional approaches is usually stated on two distinct levels of representation. I propose to use function-argument structures as fully functional logical forms of natural language sentences. To satisfy this goal, I must be able to show how they are derived from corresponding natural language sentences, how they are interpreted so as to make available only those readings allowed by the locality property, and how they are utilized to specify semantic well-formedness conditions. Two important consequences of this proposal are that for the purpose of making logical interpretation we do not need an extra level of representation such as Logical Form and we also do not need a version of rule schema such as quantifying-in that is known to be overgenerating.

The natural language sentences that are to be covered to show the point include such constructions as complex NPs, coordination of various constituents, quantifier-bound pronouns and *Wh*-phrases. For a compositional derivation of function-argument structures, I use a version of Combinatory Categorial Grammar [2]. This choice of a grammatical formalism for the purpose of this work has some interesting consequences. The semantic forms of quantified NPs are placed in argument positions not in scope-taking positions in our function-argument structures.

# Scott Prevost

Department of Computer and Information Science
prevost@linc.cis.upenn.edu

## Information Based Intonation Synthesis

Text-to-speech systems often produce intonation contours that are improper or unnatural in certain contexts, primarily due to the lack of consideration of syntactic, semantic and discourse level structures. In an attempt to produce realistic contours based on discourse context, several researchers have therefore turned their attention to meaning-to-speech systems, where decisions concerning pitch accent selection and placement are based on an underlying representation of the message to be conveyed rather than simply its surface realization (cf. Davis and Hirschberg [1] and Zacharski *et al.* [8]). Such systems have relied primarily on notions of *givenness* or *previous mention* to determine when an item should bear a pitch accent, generally accenting an item on first mention and de-accenting it thereafter. A major problem with such an approach is the inability to handle *contrastive stress*, a phenomenon that may cause an established discourse entity to bear an accent, regardless of its given status, specifically because it stands in contrast to some other salient discourse entity.

Consider a situation in which a physician inadvertently orders the wrong procedure in the treatment plan for a patient—say a left *thoracotomy* rather than the intended left *thoracostomy*. Now suppose a diagnostic critiquing system such as TraumAID verbally identifies the discrepancy between the physician's plan and the plan devised by the system, as shown below.

**a.** You seem to have confused the THORACOTOMY and THORACOSTOMY procedures in your plan for this patient.

**b.** I propose doing a left thoracostomy.

The examples below identify the four accentual possibilities for the noun phrase *a left thoracostomy* in sentence b above.

**b.** I propose doing a left THORACOSTOMY.

**b'.** I propose doing a LEFT THORACOSTOMY.

**b".** I propose doing a LEFT thoracostomy.

**b'''.** I propose doing a left thoracostomy.

Of these, only b and b' are acceptable, precisely because they highlight *thoracostomy*, which stands in direct contrast to the previously established *thoracotomy*. Unfortunately, employing previous mention heuristics for assigning the intonational contour produces the improper pattern shown in b".

The purpose of the present research is to develop a system which employs discourse structure and information structure to assign pitch accents on the basis of contrastive properties in synthesized speech motivated by simple database queries. The current implementation is able to make the types of intonational distinctions shown in the examples below.

```
Q: I know what's recommended for the PERSISTENT pneumothorax,
   but which procedure  is recommended for the SIMPLE pneumothorax?
       L+H*        L-H                            H*          L-L%

A:
   A left THORACOSTOMY  is recommended for the SIMPLE pneumothorax.
             H* L-L                            L+H*        L-H%

Q: I know what's recommended for the PERITONITIS,
   but which condition  is a left THORACOSTOMY recommended for?
       L+H*        L-H                H*                   L-L%

A:
   A left THORACOSTOMY is recommended for  the simple PNEUMOTHORAX.
             L+H*                    L-H                    H* L-L%
```

The intonation contours in these examples (shown in Pierrehumbert-style notation, see [2]), are quite different and cannot be interchanged without sounding strikingly unnatural. The two tunes (L+H* L-H and H* L-L) associate different discourse functions with the constituents over which they are distributed. In the paradigm of *wh*-queries and responses, the L+H* L-H tune seems to represent the "theme" of the utterance–what the discourse participants are talking about. The H* L-L tune, on the other hand, marks the "rheme"–what is being said about the theme. Moreover, the placement of the pitch accent (L+H* or H*) within such a tune marks the focus of the interpretation of the theme or rheme.

In the examples shown above, one can easily see that intonational phrase boundaries do not necessary correspond to traditional syntactic boundaries. Steedman, however, has previously argued that under flexible notions of syntactic constituency offered by Combinatory Categorial Grammar (CCG), syntactic and prosodic bracketing can be considered isomorphic [6]. We exploit this work in building a database query system that produces intonationally-natural spoken responses.

Input to the database query system is given textually with intonation represented symbolically using Pierrehumbert's notation. A simple CCG-parser determines a semantic representation for the query as well as interpretations for its thematic and rhematic constituents, where the semantic representation for a *wh* question is represented as an open proposition in the lambda calculus. As each question is processed, a discourse model is updated to include a representation for the current theme.

The semantic representation of the question is utilized by a *content* generation module to produce a semantic representation for the response, accomplished by instantiating the variable in the open proposition. The representations for the theme and rheme of the response are determined by mapping the rheme of the question onto the theme of the response. The focus of the rheme, which represents some discourse entity, is determined on the basis of contrastive stress by constructing a set of alternative entities from the discourse model and employing the contrastive stress algorithm described in Prevost and Steedman [4], [5].

A *CCG* generator [3] works from the output of the strategic generator to produce a string of words for the response along with appropriate intonational annotations. The generation mechanism employs a top-down "functional"-head driven scheme that utilizes the same CCG rules as the parser. The output string can then be easily converted to input for a text-to-speech system, without modifying the underlying design or algorithms of the speech synthesizer. Currently we use the Bell Laboratories TTS system to produce the spoken results.

# Ellen F. Prince

Professor, Department of Linguistics
ellen@central.cis.upenn.edu

## Discourse/pragmatics

Keywords: Discourse functions of syntax, Reference, Language contact, Yiddish

I am interested in that part of linguistic competence that underlies the use of particular linguistic forms in particular contexts, where the choice is not entailed by sentence-grammar or truth-conditional meaning. In particular, I am interested in the choice of referential expressions and syntactic constructions. I am also interested in the effects of language contact on this domain. The bulk of my research has focused on English and Yiddish.

# Lance A. Ramshaw

Postdoc, Institute for Research in Cognitive Science
Assistant Professor, Bowdoin College Computer Science Department
ramshaw@linc.cis.upenn.edu

## Learning Language Models from Corpora

Keywords: Rule sequence learning, Hidden Markov models

This effort focuses on deriving statistical models of language from either annotated or unannotated corpora. Building on earlier work with Ralph Weischedel and others at BBN [6], I have been exploring various schemes for adding probabilistic knowledge to language models. In work with Neel Smith of the Classics Department at Bowdoin College, I have built a system that assigns part-of-speech tags to Ancient Greek text from the Perseus corpus, to determine how well we could do with HMMs using fully unsupervised training (since the corpus is not annotated), or using a mix of supervised training on a small subcorpus combined with unsupervised training on the remainder. I have also applied the same techniques to a hand-tagged corpus of the Septuagint Greek version of the first five books of the Bible.

As an alternative to HMMs for statistical language modeling, Mitch Marcus and I have recently been exploring the mechanism of transformational rule sequence learning, as proposed by Eric Brill in his recent dissertation [1]. This intriguing approach makes efficient use of a minimal amount of supervised training data by combining it with rule templates that define the kinds of patterns in the data that are likely to influence the variables the model is trying to predict. The final model is encoded as a list of pattern-action rules, derived by mapping the rule templates over the training corpus. Brill's own work has shown that such rule sequence models can outperform the best traditional models for part-of-speech tagging, and that they can also be applied successfully to other tasks like building phrase structure trees.

Our recent efforts have centered on developing a better understanding of this rule sequence learning technique, based mostly on part-of-speech tagging experiments in Greek and in English, though recently we have also begun to experiment with using rule sequences to predict letter-to-sound mappings. By means of a linked implementation of the training algorithm that allows much more rapid training runs, we have been able to explore the mechanism's behavior when allowed to run to exhaustion, learning all rules that improve performance at all, rather than cutting off at some threshold.

While the results of any new mechanism for part-of-speech tagging must be compared with the performance of the standard HMM techniques, it has become clear in the course of our work that rule sequence learning can be analyzed more naturally as an adaptation of decision trees to a situation where the problem instances are interrelated. For example, in part-of-speech tagging, the solution to one ambiguity affects the predictive environment of a neighboring ambiguity. We have written a workshop paper [5] presenting that analysis of rule sequence learning in terms of decision trees, which helps to explain why the method turns out to be quite resistant to the overtraining danger that afflicts the unsupervised training of HMM models.

Since neighboring ambiguities can affect each other, it is also true that rules in the learned sequence may depend on the results of earlier rules. Our code records those dependency structures, and we are continuing to work on methods for analyzing them, since a better understanding of those dependencies may allow, for example, overtrained rules to be excised from a rule sequence or the best parts of two different rule sequences to be merged. We are also continuing to explore the ranking measure used for rule selection; there are some initial indications that a measure based on likelihood ratios may outperform Brill's "right minus wrong" measure.

## Plan-Based Discourse Modeling

Keywords: Problem-solving, Metaplans, Ill-formedness, Spelling correction

While probabilistic methods can achieve a great deal, there will always remain a pool of more complex problems whose solution depends on a deeper understanding of the text than current statistical methods can hope to achieve. This deeper level of analysis is particularly important in expert advising systems that carry on a dialogue with the user. Adequate understanding of such discourse requires modeling the user's intentions, which include both task-specific plans, discourse-related metaplans (like "Introduce Topic"), and also metaplans relating to the problem-solving process [2] (like "Explore Possible Subplan"). A three-level discourse model [3] can be useful in such settings. The resulting deeper level of understanding allows increased robustness, since even awkward "alias" errors [4] can be handled, as when a spelling error happens to result in a different, correctly-spelled but unintended word.

# Lisa F. Rau

Visiting Professor, Department of Computer and Information Science
lrau@cis.upenn.edu

# Industrial Applications of NLP

Keywords: Text clustering, Summarization and automatic extraction

I am on a one-year research sabbatical (academic year 1993-1994) at the University of Pennsylvania's Computer and Information Science's Department, sponsored by the National Science Foundation's Visiting Professorships for Women program.

During this period, I have engaged in three primary activities: 1) promoting industrial research in general and in the areas of natural language processing and text retrieval in particular, 2) furthering research in text summarization, and 3) investigating methods of automatic knowledge acquisition and clustering of text.

To promote industrial research, I gave a series of lectures on what industrial research is like to science and engineering departments across campus. Topics included the environment, compensation, job expectations and a specific comparison with a career in academia. A special colloquium on industrial applications of natural language and text retrieval was given at the Computer and Information Science Department. Trends and methods in the full-scale evaluation of real systems were detailed in the graduate seminar on the Automatic Acquisition of Information from Text. A complete review of the best examples of deployed systems in this area will appear in a forthcoming article in the Communications of the ACM [1].

In the area of text summarization, recently completed research (a collaborative effort with Ron Brandow, GE Research and Development, and Karl Mitze, Mead Data Central) investigated and evaluated methods of performing domain-independent summarization of news [3]. The ANES (Automatic News Extraction System) system constructed by GE Research and Development was compared to the Searchable Lead system developed by Mead Data Central, in the largest evaluation of automatic summarization undertaken to date.

The ANES approach utilized a combination of statistical and heuristic techniques to determine key sentences for extraction and inclusion in summaries. Statistics based on the relative frequency of terms in a document as compared to the corpus as a whole (*tf\*idf*) determined topical words. The sentences containing these words were selected using constraints that incorporated preferences such as location within a document and the presence of anaphoric references that could interfere with the readability of the final summary. The Searchable Lead technology chooses the first sentences in the news story for inclusion up to the target length for the summary. Three lengths of articles were evaluated for 250 documents by both systems, totalling 1,500 suitability judgements in all. The results of this evaluation were totally unexpected. The lead-based summaries outperformed the "intelligent" summaries significantly, achieving acceptability ratings of just over 90%, compared to the 84% acceptability of the ANES system. Future research plans are to create a hybrid approach to summarization that leverages the benefits of each method.

In the area of automatic knowledge acquisition and clustering of text, a graduate seminar (co-taught with Mitch Marcus and Paul Jacobs) reviewed the state-of-the-art in statistical and knowledge-based methods for acquiring knowledge to support natural language processing applications, for example, words senses, lexical relations, discourse relations.

Research on automatic text clustering focused on a real-life application area; scoring free-response (also called constructed response) test responses [2]. Large-scale scoring of short answer responses would be an overburdening and tedious task to perform without computer assistance. One way that automation can play a key role is by clustering responses which have a similar meaning or implication. By taking advantage of automation in this way, whole sets of responses may be judged quickly, saving time and preventing the introduction of inconsistencies.

A system was built, GE-FRST, to experiment with methods of scoring a variety of test items involving short, natural language responses. One method experimented with was fully automatic, and two others assumed some human assistance. The fully automated method did not yield viable results. The two partially automated methods were based on pattern-matching techniques and a heuristic method of assessing lexical similarity. In-depth evaluation of the system is on-going.

# Jeffrey C. Reynar

Department of Computer and Information Science
jcreynar@unagi.cis.upenn.edu

## Automatic Discovery of Topic Boundaries

Keywords: Text structuring, Discourse boundaries

In light of people's increasing reliance on electronic access to documents, the process of discovering, or imposing, document structure is becoming important for many reasons. For example, knowing the location of topic shifts may allow information retrieval to be improved, both by allowing information retrieval systems to provide users with only the relevant sections of a document and by improving accuracy; see [3] and [7]. Also, people have speculated, see [5], that segmenting text along topic boundaries may be useful for text summarization and anaphora resolution.

In [2], Halliday and Hasan demonstrate the importance of lexical items in providing coherence to a text. When a particular topic is being discussed, words germane to that topic will be used. When a new topic is begun, a shift in vocabulary will occur. Hoey [4] explains that word repetition occurs more frequently within coherent regions of a text than across topic boundaries. These ideas form the basis of an algorithm for automatically discovering topic boundaries.

In a forthcoming paper [6], I describe a method of finding discourse boundaries which is loosely based on *dotplotting*, a graphical technique described by Church in [1]. The application of this technique to text structuring uses word repetition information to divide a text into those regions determined to be most coherent by an optimization algorithm. The method has been successfully used to "discover" the document boundaries in concatenations of *Wall Street Journal* articles.

Future directions for the work include using the algorithm to discover actual topic boundaries, testing whether the segment information improves the performance of an information retrieval system and comparing the segmentations produced by the algorithm to the segmentations produced by other text structuring algorithms or the judgments of human readers. My other interests include statistical parsing, prepositional phrase attachment, text compression, information retrieval and information theory.

# Francesc Ribas i Framis

Visiting PhD student, LSI Dept., Universitat Politecnica de Catalunya
ribas@linc.cis.upenn.edu

## Learning Appropriate Selectional Restrictions from Corpora

I'm mainly interested in the automatic extraction of lexical knowledge from text. I've been concentrating on exploring the possibility of extracting selectional restrictions using statistical techniques.

In order to build a computational lexicon it seems essential to have extensive coverage of the selectional restrictions in the domain to be worked with. In recent years, some statistical methods to extract lexical relationships (among them selectional restrictions) from corpora have been proposed ( [1], [7]).

Working in this framework, I've been doing some experiments on extracting selectional restrictions at an appropriate level of abstraction from phrasally analyzed corpora ( [8], [9]). The technique induces the relevant lexical relationships that hold between verbs and classes of nouns, using a measure of association based on the statistical relevance of the co-occurrences of verbs and nouns in a given syntactic relationship in the corpus. Using this measure, and a broad semantic taxonomy [6], a search process tries to find out the best classes on this taxonomy to convey the selectional restrictions.

Several experiments using a fragment of the Penn Treebank [5] have been carried out using different variants of the measure of association (Relative Entropy [2] and Likelihood Ratio [4]). The results have been tested against a manually built semantic concordance. Some problems not solved by the methodology have been analyzed in the light of these experiments: the impact of noise introduced by wrong noun senses, mixing of different verb senses (predicates), and finally mixing of arguments. I'm now concentrated on exploring some solutions to these problems, mainly to the noun sense-ambiguity problem, which I want to tackle by starting from some sense-disambiguated data and afterwards applying a variant of the EM algorithm [3] on a non-supervised corpus, in order to get a better estimate of the involved probability distributions. I'm also trying to define different evaluation measures to compare the selectional restrictions acquired by the different techniques explored.

# James Rogers

Postdoc, Institute for Research in Cognitive Science
jrogers@linc.cis.upenn.edu

## Logic and the Structure of Language

My research tends to involve the application of results and techniques of mathematical logic to issues involving the structure of languages. Currently, there are three areas that are most active.

### Descriptive Complexity and Language Complexity

Descriptive complexity results generally establish characterizations of *computational complexity* classes (PTIME and NPTIME, for instance) in terms of definability in particular logical languages. We are interested in such results for *language complexity* classes (the Chomsky Hierarchy, for instance). These classes are determined by the *nature* of the resources required to parse or recognize languages rather than the *quantity* of those resources. In an early result of this type, Büchi [2] showed that the class of regular languages was exactly the class of finite strings definable in *S1S*, the monadic second-order theory of successor and less-than. More recently (originally due to Doner in the late 60's [3]) it has been shown that the sets of finite trees definable in *SnS*, the monadic second-order theory of multiple successor functions, are exactly the *recognizable sets*. These are essentially the sets of derivation trees generated by Context-Free Grammars. Thus languages are strongly Context-Free iff the sets of trees analyzing the structure of their strings are definable in SnS.

This result is an extremely effective tool for establishing language complexity results for formalisms that specify languages in terms of constraints on the structure of their strings. We have used it, for example, in showing that the language licensed by a particular Government and Binding Theory account of English is strongly Context-Free [6]. Since previously it has been quite difficult to establish that such languages are even recursive, this gives an indication of the power of the approach. On the other hand, one does not expect to be able to get such results for natural languages in general (for a full GB account of Universal Grammar, for instance) at the level of Context-Free languages. Consequently, we are looking to develop similar results for larger complexity classes, and in particular for classes falling properly between the classes of Context-Free and Context-Sensitive languages.

### Formal Foundations of Grammatical Structures

Many theories of syntax analyze the structure of sentences as trees. There has been a growing body of work reasoning formally about these structures. With the goal of providing a rigorous foundation for such reasoning, we have developed an axiomatization of the first-order, and even the universal fragment of the monadic second-order, theory of finite trees of the sort typical of linguistic theories [1]. We are currently working to extend this to the theory of infinite trees, and to use this as a framework with which to relate the various logical languages which have been proposed for reasoning about trees.

In a slightly less theoretical direction, we have been working to characterize tree adjunction as a logical operation. This work has the prospect of providing a unified foundation from which to study the range of variations and extensions of the basic adjunction operation [5].

**TAGs that Generate CF Languages**

Previously we have characterized the class of pure TAGs that generate recognizable sets in terms of the structure of their sets of elementary trees [4]. We are looking to extend these results in two ways. One is to explore characterizations of the (strictly larger) class of TAGs that generate (weakly) Context-Free string languages. The other is to extend this characterization to TAGs with atomic features. This would allow these results to be applied to the current XTAG grammar. A cursory look at the predecessor of this grammar suggests that it may well generate a strongly Context-Free language. The extended result would allow us to identify in what ways, if any, the XTAG grammar exceeds the power of CFGs.

# Bernhard Rohrbacher

Postdoc, Institute for Research in Cognitive Science
bwr@linc.cis.upenn.edu

## Inflectional Morphology and the Parametrization of Heads and Specifiers

Keywords: Syntax, Acquisition, Verb Movement, pro-Drop

Chomsky suggests that "parameters of UG relate not to the computational system, but only to the [lexical entries of functional elements]" [1]. It is often assumed that abstract syntactic properties of functional elements determine parametrization. In contrast, my working hypothesis is that only overt morphological properties of functional elements are relevant for parametrization. I have argued that the highest inflectional head AgrS contains affixal material in underlying representation in exactly those languages where overt subject-verb agreement distinctively marks the person features and that D-structural AgrS-affixes are responsible for V-to-AgrS Raising (via Lasnik's Filter) and pro-drop (via Economy of Projection) [3]. This conclusion was reached on the basis of synchronic and diachronic adult data, in particular from the Germanic languages. In ongoing research with Anne Vainikka, I investigate whether child data lead to the same conclusion. Initial results concerning the acquisition of the pro-drop parameter are promising. As predicted by Economy of Projection, English [2] and German [4] children leave out subjects only when no agreement morphology is present. The situation with respect to V-to-AgrS raising is more complex. German children master verb placement before the relevant agreement morphology is acquired [4], but since adult German employs V-to-C raising, it is yet unclear what to conclude from this fact. It will be especially interesting to compare child German with child Swedish (which unlike German lacks V-to-AgrS) and child Yiddish (which unlike German has AgrS on the left). Access to child Yiddish data will be made possible by Ellen Prince. This aspect of my research is particularly exciting since virtually no work on the acquisition of Yiddish has been done.

### Other Research

With Tilman Becker, I am working on a Principles-and-Parameters analysis of Parasitic Gaps in German and a formalization of this analysis in the TAG-framework. German Parasitic Gaps are of special interest because there wellformedness is governed by overt instead of abstract case-marking, a fact that might require a rethinking of the relation between overt and abstract case within the Principles-and-Parameters approach.

# Joseph Rosenzweig

Department of Computer and Information Science
josephr@linc.cis.upenn.edu

## Instrument action specifications in natural-language instructions

Natural-language instructions often underspecify the tasks they describe. In particular, instruments which are needed to carry out a task may be left unmentioned, or, if they are mentioned, their exact role in the task to be performed may require elaboration. Information describing the performance of a task may also be distributed across several sentences, so that an analysis of discourse coherence is required to link up the mention of an instrument with the action in which it participates. In this research, instruments include tools, artifacts, naturally-occurring forces and materials, and the body parts of (human or robot) agents. These all comprise resources which an agent may have to rely on to complete a task.

I have assembled a corpus of instruction text, including over 2,000 tokens of the verb "use", to study how information about instruments is encoded in instructions. When instruments appear as the object of "use", the reader must draw more heavily on knowledge about what roles the tools usually play in tasks, rather than knowledge about the verb. To exploit this knowledge, an instruction-understanding system such as that used in the AnimNL project must have a rich representation of the semantics of nouns.

To understand how tools fit into the tasks described by verbs, the system also needs to be able to compose and decompose verb and noun meanings. I have been analyzing approximately 800 denominal verbs which transparently incorporate nouns. The relationship between the meaning of the incorporated noun and the meaning of the verb as a whole sheds light on the semantic structures involved. Meanwhile, compound nominals such as "fuel tank access panel leak check" can encode a large portion of the information about a task without a verb. In the Treebank corpus, over 24 percent of noun tokens occur within compounds. I am therefore also concerned with the semantics of noun-noun composition as it relates to the description of instruments and the understanding of their use. In addition, I am interested in methods of acquiring lexical information from corpora and machine-readable dictionaries, which can guide the choice of representations by suggesting what kinds of terminological knowledge must be accommodated.

# Deborah Rossen-Knill

Institute for Research in Cognitive Science
drossen@linc.cis.upenn.edu

## Toward a Pragmatics of Dialogue in Fiction

My thesis [15] develops and applies a pragmatic model of dialogue in fiction, an area which has received scant attention despite scholars' and writers' recognition of its unique ability to engage the reader in the story and so bring that story to life. Working on the belief that what and how a character communicates reflects his/her identity as an individual and as a group member, I analyze fictional dialogue, using three types of pragmatic theories. A modification of Austin's [1] and Searle's [16] [17] speech act theories explains the mechanics of meta-conversational comments, communication breakdowns, and techniques for creating and manipulating the fictional world. Labov and Fanshel's pragmatic theory, altered to suit particular works of fiction, reveals the character traits and interactions responsible for meta-conversational criticisms and communication breakdowns. Finally, drawing from several theories of politeness (Lakoff 1973; Brown and Levinson [3]; Leech [11], Brown and Gilman [4]) and deference (Fraser and Nolan [6]), I employ a version of politeness which emphasizes communication as a means to enhance group solidarity with minimal sacrifice of individual interests. This theory directs the reader's attention to contextual features motivating a particular speaking strategy, which in turn reveals the character's position in and relationship to the group and the speech-event [2].

Physically imagined, the model has at its core two characters, each having at least a representation as an individual and a representation as a group member. Surrounding, motivating and shaping characters' conversations are contexts, each of which has corresponding conventions. My model distinguishes between three kinds of context: that which is external to an individual, that located in the individual mind, and that which delimits the boundaries of the fictional world. I view external contexts as frames [7] which constrain the conversation according to the relationship between conversants and the objects or information being discussed, as well as the participants' individual and collective goals. The internal context involves an individual's mental model of his/her world [18]. The fictional context involves the world described by the fictional text; it both isolates and gives a sense of "reality" to the characters. All boundaries of contexts are permeable, often making the description of any one context dependent on another. When applied to a work of fiction, the model systematically guides the analysis of dialogue in order to explain and interpret character interactions.

An abridged version of chapters 1 and 2 appears as [13].

## The Pragmatics of Verbal Parody

This paper [12] is joint work with Richard M. Henry.

We argue that verbal parody involves a highly situated, intentional and conventional expressive made up of four essential acts: 1) the intentional representation of the object of parody (related to quotation [5], 2) the flaunting of the verbal re-presentation, 3) the critical act, and 4) the comic act. To successfully create a verbal parody, a speaker must manipulate all four essential acts with the intent to communicate parody. We also address the potential scope of parody in real communication. We argue that the object of parody may be anything in the world; that a single parodic act may have multiple objects; and that the re-presenting verbal expression of the parodic speech act may function as a direct or indirect non-parodic speech act, which may be enhanced or inhibited by the parody. We also explain how parody serves to celebrate the object it apparently ridicules, by appealing to politeness theory [3].

**_The Princess Bride_ and the Parodic Impulse: The Seduction of "Cinderella"**

This paper [8] is joint work with Richard M. Henry.

We examine the popular film _The Princess Bride_ to show how screenwriter William Goldman and director Rob Reiner use parody to simultaneously reject and reaffirm the values of fairy tale True Love. We extend here our model of parodic communication to offer a particularly overt demonstration of its workings: _The Princess Bride_ "re-presents" earlier depictions of fairy tale True Love, "flaunts" that re-presentation, and so "critiques" True Love in a "comical" way. Whereas oftentimes parody leaves its audience with simultaneous messages of critique and humor, _The Princess Bride_ pays tribute to the very values it apparently rejects. The tribute to True Love is a function of the interactions among the four parodic acts: the heightening of the comic act couples with the muting of the critical act weakens the critique; and the heightening of the re-presenting act reaffirms the values under critique. The filmmakers manipulate these interactions to advance two simultaneous messages: "'Twue Rove' is funny," and "True Love is good."

# Anoop Sarkar

Department of Computer and Information Science
anoop@linc.cis.upenn.edu

## Incremental LR Parsing of TAGs

Keywords: Parsing, Tree Adjoining Grammar

The construction of deterministic bottom-up left to right parsing of Tree Adjoining Languages (TALs) is described in [3] [4]. This work extends their approach to handle the incremental parsing of TALs.

Parser generation provides a fast solution to the parsing of input sentences as certain information about the grammar is precompiled and available while parsing. However, if the grammar used to generate the parser is not completely fixed and needs frequent modification then the time needed to parse the input is determined by both the parser and the parser generator.

The main application area for Tree Adjoining Grammars (TAGs) has been the linguistic description of natural languages. In such an area grammars are usually large and very rarely static, modifications to the original grammar are commonplace. In such an interactive environment, conventional LR-type parsing suffers disadvantages which we avoid by allowing the incremental incorporation of modifications to the grammar in a LR-type parser for TALs. This paper extends the work done on the incremental modification of LR(0) parser generators for CFGs in [1] [2]. We define a lazy and incremental parser generator having the following characteristics:

- The parser is generated in a lazy fashion, *by need* while parsing the input. The parser generator now does not have the overhead of computing the parse table for the entire grammar.

- The parser generator is incremental. Changes in the grammar trigger a corresponding change in the already generated parser. Parts of the parser not affected by the modifications in the grammar are reused.

- Once the needed parts of the parser have been generated, the parsing process is as efficient as a conventionally generated one.

# B. Srinivas

Department of Computer and Information Science
srini@linc.cis.upenn.edu

## Disambiguation of Supertags – Almost Parsing

This work is jointly being pursued with Aravind K. Joshi.

Part-of-speech disambiguation techniques (*taggers*) are often used to eliminate (or substantially reduce) the part-of-speech ambiguity prior to parsing. The taggers are all local in the sense that they use information from a limited context in deciding which tag(s) to choose for each word. As is well known, these taggers are quite successful.

In a lexicalized grammar such as the Lexicalized Tree-Adjoining Grammar (LTAG), each lexical item is associated with at least one elementary structure (tree). The elementary structures of LTAG localize dependencies, including long distance dependencies, by requiring that all and only the dependent elements be present within the same structure. As a result of this localization, a lexical item may be (and, in general, almost always is) associated with more than one elementary structure. We will call these elementary structures *supertags*. Note that even when a word has a unique standard part-of-speech, say a verb (V), there will in general be more than one supertag associated with this word. Since when the parse is complete, there is only one supertag for each word (assuming there is no global ambiguity), an LTAG parser [8] needs to first search a large space of supertags to select the right one for each word before combining them for the parse of a sentence.

Since LTAGs are lexicalized, we are presented with a novel opportunity to eliminate or substantially reduce the supertag assignment ambiguity by using local information such as local lexical dependencies, prior to parsing. As in the standard parts-of-speech disambiguation, we can use local statistical information, in the form of n-gram models based on the distribution of supertags in a LTAG parsed corpus. Moreover, since the supertags encode dependency information, we can also use information about the distribution of distances of the dependent supertags for a given supertag.

Note that as in the standard part-of-speech disambiguation, the supertag disambiguation could have been done by a parser. However, carrying out the part-of-speech disambiguation prior to parsing makes the job of the parser much easier and therefore faster. Supertag disambiguation as proposed in this research reduces the work of the parser even further. After supertag disambiguation, we have effectively completed the parse and the parser need 'only' combine the individual structures; hence the name—*almost parsing*. This method can also serve to parse sentence in cases when the supertag sequence after the disambiguation may not combine to form a single structure.

The main goal of this work is to investigate techniques for disambiguating supertags, and to evaluate their performance and their impact on LTAG parsing. Experiments using the n-gram and dependency models for supertag disambiguation are underway using the LTAG parsed corpus of Wall Street Journal sentences. The preliminary results [6] are promising. Although investigated with respect to LTAG, these techniques are applicable to lexicalized grammars in general. These techniques are being tried out for the wide-coverage CCG [2] that is being bootstrapped from the XTAG system [4].

# Evaluating a Wide-Coverage Lexicalized Grammar

Keywords: XTAG, Parser Evaluation, Wide-Coverage Grammars, Statistical Parser

This work has been jointly pursued with Christine Doran, Seth Kulick and Anoop Sarkar.

In this work [3] we present the preliminary results of the performance evaluation of the wide-coverage, non-stochastic English grammar of the XTAG system [4] and compare it against the performance of the IBM statistical parser [5] and the Alvey NL Tool parser [1].

XTAG has recently been used to parse Wall Street Journal[1], IBM manual, and ATIS corpora as a means of evaluating the coverage and correctness of XTAG parses. Preliminary results are shown in the Table 1. For this evaluation, a sentence is considered to have parsed correctly if XTAG produces parse trees.[2]

| Corpus | # of Sentences | % Parsed | Av. # of parses/sent |
|---|---|---|---|
| WSJ | 6364 | 39.09% | 7.53 |
| IBM Manual | 1611 | 75.42% | 6.14 |
| ATIS | 649 | 74.42% | 6.0 |

Table 1: Performance of XTAG on various corpora

A more detailed experiment to measure the crossing bracket accuracy of the XTAG-parsed IBM-manual sentences has been performed. Of the 1600 IBM sentences we have parsed (those available from the Penn Treebank [7]), only 67 overlapped with the IBM-manual treebank that was bracketed by University of Lancaster.[3] The XTAG-parses for these 67 sentences were compared[4] with the Lancaster IBM-manual treebank.

| System | # of sentences | Crossing Brackets | Recall | Precision |
|---|---|---|---|---|
| XTAG | 67 | 80% | 84.32% | 59.28% |
| IBM Statistical parser | 1000 | 86.2% | Not Available | Not Available |

Table 2: Performance of XTAG on IBM-manual sentences

We also compared XTAG to the Alvey Natural Language Tools (ANLT) Parser, and found that the two performed comparably. We parsed the same set of LDOCE Noun Phrases as presented in Appendix B of the technical report [1] using the XTAG parser. In this experiment, we have compared the total number of derivations obtained from XTAG with that obtained from the ANLT parser. Table 3 summarizes the results of this experiment.

Further work is being done to compare the performance of XTAG on the Lancaster-treebank of IBM sentences. Also we are interested in getting more data of the kind available in [1], so that we can make further comparisons with the Alvey parser.

---

[1] Sentences of length < 16 words.

[2] Verifying the presence of the correct parse among the parses generated is done manually at present.

[3] The treebank was obtained through Salim Roukos (roukos@watson.ibm.com) at IBM.

[4] We used the parseval program written by Phil Harison (phil@atc.boeing.com).

| System | # of NPs | # parsed | % parsed | Maximum derivations | Average derivations |
|--------|----------|----------|----------|---------------------|---------------------|
| ANLT Parser | 143 | 127 | 88.81% | 32 | 4.57 |
| XTAG Parser | 143 | 124 | 86.71% | 28 | 4.14 |

Table 3: Comparison of XTAG and ANLT Parser

# Mark Steedman

Associate Professor, Department of Computer and Information Science
steedman@linc.cis.upenn.edu

## Combinators and Grammars for Natural Language Understanding

My research interests cover a range of issues in the areas of computational linguistics, artificial intelligence, computer science and cognitive science, including syntax and semantics of natural languages and programming languages, parsing and comprehension of natural language discourse by humans and by machine, natural language generation, and intonation in spoken discourse. I also occasionally work on formal models of musical comprehension.

Much of my research since completing my graduate work has been on two problems in computational linguistics. The first concerns a theory of natural language syntax and its relation to "incremental" syntactic and semantic processing of spoken and written language. The research demonstrates a direct relation between certain problematic natural language constructions and certain purely local, variable-free, combinatory operations on functions, such as functional composition. The constructions in question involve unbounded dependencies between syntactic elements, such as those found in relative clauses and in coordinate constructions. The combinatory operations are related to some of the simplest combinators which have been used to provide a foundation for applicative systems, including the $\lambda$-calculus and the related programming languages. The research addresses a number of questions of practical importance. The weaknesses of most current theories of grammar in the face of the full range of coordination phenomena means that existing computational grammars have the characteristics of unstructured programs—that is, they are non-modular and hard to modify, placing practical limitations on the size and portability of the systems that include them. The standard theories show a similarly bad fit to a number of other phenomena of practical importance, notably phrasal prosody and intonation. Most of my current work is in this latter area, in particular in the problem of synthesising contextually appropriate intonation in limited conversational domains. Some of this work is being pursued in collaboration with the Graphics Laboratory, and concerns animating conversation between autonomous agents whose utterances, including intonation and gesture, are specified entirely autonomously and by rule.

My second principal research interest concerns a computationally-based semantics for tense and temporal reference, and exploits the advantages of computational models for capturing phenomena which are presupposition-laden and involve interactions with non-sentence-internal knowledge. The work shows that the primitives involved in this domain are not solely (or even primarily) temporal, but rather are concerned with "contingent" relations between events, such as causation. This project also addresses a practical concern, for any database that is to be interrogated or updated in natural language making use of tense and related categories is certain to to require structuring in the same way. A number of domains are under investigation, including certain problems in the graphical animation of action sequences.

# Matthew Stone

Department of Computer and Information Science
matthew@linc.cis.upenn.edu

## NL Generation for Task-Oriented Dialogue

Keywords: Generation, Dialogue, Semantics, Gesture, Intonation, Facial expressions

As part of the Gesture Jack project, I have worked on the development of prototypes for conversational agents. The agents we built construct plans, verify the appropriateness of these plans with their conversational partners, obtain their partner's assistance and cooperate to carry out adopted plans. All of the dialogue that the agents use is automatically generated; the output specifies appropriate gestures and intonation in addition to text. See [1] and [2].

In the prototype, the agents maintained a relatively detailed representation of their dialogue, so that the algorithms of Prevost and Steedman [3] could be used to determine appropriate intonation. The placement of gesture was also controlled by discourse and semantic status of the material to be conveyed. To be precise, a gesture was produced for a concept when three conditions were met: first, the concept was part of a description of an entity or action new to the discourse; second, the entities or actions were introduced as part of the rheme of an utterance (the constituent with the most significant new information for the hearer); and third, a spatial (or spatializable) representation of the concept was stored in the lexicon.

The most difficult part of the prototype came in designing the agents to ask natural-sounding questions. In our domain, English questions often present a collection of constraints on an unknown entity, all of which are needed for the entity to play a successful role in a plan. Accumulating these multiple constraints requires reasoning about unknown, hypothetical entities during planning, and then using linguistic principles to organize the resulting constraints into a coherent sequence. A detailed description and justification of the procedure our agents used can be found in [5].

My current work concerns elaborating the Gesture Jack model to include principled algorithms for automatically generating appropriate facial expressions—according to *appraisals* of how ongoing communication impacts the plans and goals of the agents about themselves and their conversation— and modal verbs—in terms of the hypotheses and reasoning required in making plans. Preliminary work on a particular modal, epistemic *must*, is reported in [4].

# Anne Vainikka

Institute for Research in Cognitive Science
vainikka@linc.cis.upenn.edu

## The layered syntactic tree

Most of my work concerns the nature of the syntactic tree in natural language. The following open questions guide my research: What is the basic architecture of the syntactic tree? Is it two-dimensional, or do three-dimensional trees exist? Is it constructed derivationally, or is it a more static (representational) structure? What do morphosyntactic processes such as case marking and agreement tell us about the tree? What is the status of functional projections in the tree? Are they the same in all languages? If not, how do they differ, and why? How much of parameter setting derives from the architecture of the tree?

There appears to be a one-to-one correlation between positions in the tree and morphosyntactic markers such as case and agreement (see e.g. my work on Finnish [5] [7] [8], Icelandic [4], Warlpiri, Dyirbal [6], and Quiche). If this is correct, it provides us with a handle for isolating syntactic positions in a particular grammar, based on the distribution of observable, overt morphosyntax. Data from first language acquisition of case and other functional elements in English [9], German, Finnish [2], Tamil [3] and Mauritian Creole [1] support this view. For example, English-speaking children acquire nominative case (associated with the Spec(IP) position) and inflectional elements on the verb (associated with INFL) at the same time, suggesting that the IP projection develops at this point [9].

Furthermore, the acquisition data viewed in this fashion reveal a pattern of development corresponding to the bottom-up development of the syntactic tree, from VP to IP to CP, and from NP to DP. We have also argued for a similar gradual development of the syntactic tree in adult *second* language acquisition (acquisition of German by Turkish, Korean, Spanish and Italian speakers [10] [12] [11]); we are in the process of attempting to define exactly how to characterize the *differences* in L1 vs. L2 acquisition in these terms.

# Bonnie Lynn Webber

Professor, Department of Computer and Information Science
bonnie@central.cis.upenn.edu

For many years, the focus of my research was on how we use language to get information from each other, when we may not either know exactly what we need or be able to predict the exact results of our communicative actions. I felt that the results of this research would help us to design information systems that could accommodate the real behavior of information seekers and providers. The work done with colleagues (Joshi, Finin, Weischedel) and students (Mays, McCoy, Pollack, Hirschberg, Cheikes) on topics related to "Cooperative Responses" and "Expert Questions" had this as their primary goal.

More recently, in connection with work on model-based human figure animation and on medical decision-making (see CliffNotes entry for TraumAID), I have become interested in some related problems: (1) how we use language to direct or guide the behavior of other agents, and (2) how we use physical behavior to get information. Both underlie research I am supervising, as described below.

## Understanding Instructions

Keywords: Understanding Instructions, Human Figure Animation

The work I am supervising on instruction understanding takes the view that instructions are *texts* intended to be understood *in context*, produced by an instructor with *more experience* than the instructee. (1) That instructions are texts means they rely on an interaction of language, world knowledge and reasoning to get their message across, and do not of themselves suffice to inform an agent of what to do or what to expect. (2) That instructions are meant to be understood in context means that an agent's understanding of a text evolves over time: that while some degree of understanding is needed to make their content available at the right point, full explication only comes through situated execution. (3) That instructors have more experience means that their words are worth trusting to some extent, even if the world initially provides no corroborating evidence.

There are many interesting questions that arise in the context of instruction understanding. One such topic is how and when instructions convey the coupling between perception and action needed to successfully carry out tasks. For example, the instruction "Wait until the paint has dried", relies on the agent's knowing already that s/he must interleave waiting with one or more types of perceptual tests. The fact that the agent can't rely on the paint itself to tell him when it's dry is not apparent from the instruction alone. On the other hand, instructions such as "Go to the first full intersection. There is a set of lights there, and the cross street is Diamond Hill." and "Go on Rte. 78 until you reach the exits for the Garden State Parkway. It is about 12 miles." convey what the agent can either monitor for or use as additional evidence in behavioral decisions. Another topic is how agents who know how to perform an action when it involves a single object (e.g., carrying a box, washing a dish, etc.) use that knowledge and their awareness of the current situation to understand and respond to instructions to perform the action on multiple objects. It is clear that they don't merely iterate the same action on each object: rather, they may multiplex in a variety of different ways.

There are many more issues that come up in the context of language and behavior, many of which are ripe for investigation. (See CliffNote entries for AnimNL, Christopher Geib, Libby Levison, Michael B. Moore, and Joseph Rosenzweig.)

## Decision Support for Multiple Trauma Management

Keywords: Clinical decision support systems, Diagnosis, Planning, Critiquing, Physiological Modelling, Anatomical Reasoning

For the past eight years, I have been working with John R. Clarke, M.D. (Dept. Surgery, Medical College of Pennsylvania) and our students to develop a clinically-viable computer-based system that can provide valuable support for physicians dealing with trauma patients. Our original intention was that the system, now called TraumAID, would serve to provide on-line advice, telling the physician what clinical procedures were currently called for and why they were needed. A subsequent experiment with an putting an earlier version of TraumAID in the MCP Emergency Room, changed our opinion, and our current concept is better characterized under the title "Real-Time Quality Assurance". For RTQA, TraumAID's role is to evaluate the physician's orders and verify that they are compatible with TraumAID's understanding of the case. If they diverge too greatly from the standards of cost-effective care embodied in TraumAID, then TraumAID must deliver a critique of those orders.

TraumAID and its evolution remain of great interest to me, not only because of the potential good it can provide but also because of the parallels between reasoning, planning and acting in clinical management, and the same activities carried out in Natural Language interaction. Work on each informs the other, to the greater enrichment of both.

(See CliFF Note entries for TraumAID, Carberry, Gertner, and Kaye for further information on this work.)

Quality Control During the Initial Definitive Management of Trauma Patients. Initial Trauma Management Plans. Submitted to the Annual Meeting of the American Association for the Surgery of Trauma, September, 1993.

## Michael White

Department of Computer and Information Science
mwhite@linc.cis.upenn.edu

## A Computational Approach to Aspectual Composition

Language provides us with expressions for describing *processes* that may go on for indeterminate periods, as in *Jack poured water into the container*; it also provides us with expressions for describing *events* with determinate endpoints, as in *Jack filled the container with water*. This distinction is most evident when such expressions are combined with measure adverbials, like *for ten seconds*. In the case of process expressions, measure phrases are unproblematic — witness *Jack poured water into the container for ten seconds*. In the case of event expressions, however, measure phrases are ill-formed under single event readings — witness *\*Jack filled the container with water for ten seconds*.

As we know from the works of Dowty [1], Verkuyl [9] and many others, language is not so inflexible as to localize this distinction to the verb alone, contra Vendler [8]. Given different arguments, such as a direct object with an amount phrase, a verb like *pour* yields an event expression: *Jack poured twenty liters of water into the container*. This interaction is the problem of *aspectual composition*. Contrariwise, given modifiers such as the progressive, a verb like *fill* yields an expression useful for describing an ongoing process: *Jack was filling the container with water*. Curiously, such expressions may be employed without entailing that the endpoints are ever actually reached—note that the preceding sentence may be true even if Jack is interrupted and never finishes filling the container. This phenomenon is known as the *imperfective paradox*.

In recent years the linguistics and philosophy literature has seen much discussion of the intuitive idea that events and processes are ontologically distinct entities, on a par with objects and substances. One might expect that an adequate formalization of these entities would directly yield an explanation of the problem of aspectual composition and the closely related imperfective paradox. However, no such account has been developed to date. While Hinrichs Hinrichs [2] and Krifka [6] both make use of these distinctions, they do not base their accounts of aspectual composition directly upon them; moreover, though Jackendoff [4] does suggest such an approach, his account is not sufficiently formalized to make precise predictions. In the first part of my thesis [13], I investigate to what extent the parallel ontological distinctions introduced above, if properly understood, may be directly employed in an explanatory formal account of these two problems.

The results of my investigation are promising. By employing a *categorial grammar* to translate English sentences into an *order-sorted logic*, the account is able to explain why *Jack ran two miles along the river* is logically equivalent to *Jack ran along the river for two miles*, while at the same time explaining why *\*Jack ran to the bridge for two miles* is ungrammatical, and thus obviously not equivalent to *Jack ran two miles to the bridge*. Furthermore, the account is the first one to explain not only why *Jack ran for ten minutes* entails *Jack ran some distance in ten minutes*, but also why different temporal adverbials are appropriate in the two cases. Naturally, the account also covers much of the more well-known data.

In the second part of my thesis [13], I investigate some of the computational implications of the linguistic account, by way of two case studies. In the first one, I show how the *event calculus* of Kowalski and Sergot [5] may be extended to cover many of the desired inferences (cf. also [12]). Following Moens [7], I then show how the calculus facilitates the implementation of a simple statement verifier which allows for a much greater range of natural language queries than is usually the case with temporal databases. Finally, I conclude by discussing how this prototype system could be made more effective and efficient.

In the second case study, I show how the *interpretation as abduction* approach advocated by Hobbs and his colleagues [3] can be augmented with some basic constraint logic programming techniques to build a system which produces simple animations from natural language descriptions (cf. also [10] and [11]). The study focuses on the relevance of the linguistic account to discourse interpretation and spatio-temporal semantics. Particular attention is paid to the spatial prepositions *along, towards,* and *away from*, which do not determine endpoints, as well as to the interaction of event descriptions and object reference.

# David Yarowsky

Department of Computer and Information Science
yarowsky@unagi.cis.upenn.edu

## Statistical Techniques for Lexical Ambiguity Resolution

The focus of my research has been the development of several techniques for resolving lexical ambiguity. Recently I have proposed a general-purpose statistical decision procedure for lexical ambiguity resolution using a variant of decision lists [7]. The algorithm considers multiple types of evidence in the context of an ambiguous word, exploiting both local syntactic patterns and more distant word associations to generate an efficient and perspicuous recipe for resolving a given ambiguity. The algorithm achieves considerable advantages by basing its classifications solely on the single most reliable piece of evidence identified in the target context. It effectively combines the strengths of both n-gram taggers and Bayesian classifiers and is useful for both syntactic and semantic ambiguities [12] [13].

I have applied this algorithm to a set of varied but fundamentally similar problems: word sense disambiguation, lexical choice in machine translation, homograph disambiguation in speech synthesis, capitalization restoration, and accent restoration in languages such as Spanish and French [8] [10] [11] [12] [14]. This last example has substantial advantages as a case study in that it provides an objective ground truth for automatic evaluation, involves both semantic and syntactic ambiguities, and is an important problem in multi-lingual language processing. I have also tested this approach on artificial sense distinctions called *pseudo-words*, as well as lexical distinctions likely to arise in optical character recognition, speech recognition, and spelling correction.

A primary reason why decision lists are as effective as they are for these tasks is the observation that a word exhibits only one "sense" per collocation. I empirically explored the nature and reliability of this property in [10]. This distributional tendency may be utilized in conjunction with the related hypothesis of only "one sense per discourse" [3].

In earlier work with Bill Gale and Ken Church I developed Bayesian classifiers for word sense discrimination using parallel bilingual corpora as training data [1] [2] [4] [5]. We applied similar techniques to the problem of authorship identification and used this approach in conjunction with the EM algorithm to label proper names in text as person or place [6].

I independently proposed and developed a class-based approach to sense disambiguation [9] using statistical models of the typical context of words in Roget's thesaurus classes such as ANIMALS or MACHINERY. The sense of a polysemous word such as crane was determined by calculating the most likely category given surrounding context. By exploiting the monosemy of most words and robustly handling the noise cause by polysemous ones, this approach overcame the need for hand-tagged training data, and achieved 92% mean accuracy disambiguating a set of polysemous words in raw, untagged text.

In support of the above research, I have developed a number of tools for corpus analysis. These include several stages of text markup, corpus encoding, hierarchical indexing, and a set of programs that do effecient, parallel search for diverse types of linguistic patterns in a 450 million word text database.

My research interests also include the statistical induction of natural language morphology. I've developed several machine learning techniques that generalize from lists of root-form/inflected-form pairs to correctly analyze new examples. I am currently exploring the problem of inducing morphology from unannotated wordlists, identifying a set of suffixes, spelling change rules and root/inflected-form pairs that provide a minimum-description-length analysis of the available data in several languages.

# Part III
# Projects and Working Groups

# The AnimNL Project

Department of Computer and Information Science

Norman Badler, Bonnie Lynn Webber, Mark Steedman, Brett Achorn, Welton Becket, Barbara Di Eugenio, Christopher Geib, Libby Levison, I. Dan Melamed, Michael Moore, Joseph Rosenzweig, Michael White, Xinmin Zhao.

The AnimNL project (for "Animation and Natural Language") aims to enable people to use Natural Language instructions to tell animated human figures what to do. Potential applications include not only human factors analysis (creating animated task simulations from instructions) but also small group training (enabling people to collaborate through language with virtual human agents in virtual environments).

Our work has focussed on procedural instructions and warnings, such as those found packaged in with appliances and equipment and in the pages of "how to" books—for example,

- Depress door release button to open door and expose paper bag.

- Unplug the vacuum cleaner if you leave the room.

(both from the *Royal* CAN VAC<sup>TM</sup> Owner's Manual). Such instructions assume that agents may be new to the tasks specified, but that they have both the experience and world knowledge needed to understand the instructions, and the skills needed to carry them out.

Besides its potential applications, AnimNL provides a rich framework in which to analyse the semantics and pragmatics of instructions, and to characterize how understanding evolves through activity. The latter has not been studied systematically before, but is especially important for linking language with behavior. The link requires that language understanding no longer be viewed merely as "front-end processing". What an agent takes an instruction to mean must be able to evolve as the agent acts. Two examples should suffice:

- When an agent is told to *Go into the kitchen to get the coffee urn*, he does not need to ground the definite expression *the coffee urn* before he begins to act. All that is required is that he be able to establish a referent once he gets to the kitchen. The understanding process must be able to allow for this delay.

- When an agent is told *Vacuum against the direction of the pile to leave it raised*, the agent can find out through vacuuming what direction of sweep leaves the carpet pile raised. Again, he does not need to know the referent before starting to act, but he must be able to use the instruction to guide what it is he needs to know.

AnimNL builds upon the *Jack*<sup>TM</sup> animation system developed at the University of Pennsylvania's Computer Graphics Research Laboratory. Animation follows from model-based simulation. *Jack* provides biomechanically reasonable and anthropometrically-scaled human models and a growing repertoire of behaviors such as walking, stepping, looking, reaching, turning, grasping, strength-based lifting, and collision-avoidance posture planning [1]. Each of these behaviors is environmentally reactive in the sense that incremental computations during simulation are able to adjust an agent's performance to the situation *without further involvement of the higher level processes* [2] unless an exceptional failure condition is signaled. Different spatial environments can easily be constructed and modified, to enable designers to vary the situations in which the figures are acting.

Trying to make a human figure move in ways that people *expect* a human to move in carrying out a task is a formidable problem: human models in *Jack* are highly articulated, with over 100 degrees of freedom [1]. While the environment through which a *Jack* agent moves influences its low-level responses, we have found that a great many behavioral constraints can be derived through instruction understanding and planning. Further descriptions of work being done in the context of AnimNL can be found in CLiFF Note entries for Geib, Levison, Moore, Webber and White.

# The Gesture Jack project

Department of Computer and Information Science
justine@central.cis.upenn.edu

Justine Cassell, Norm Badler, Mark Steedman, Catherine Pelachaud, Brett Achorn, Tripp Becket,
Brett Douville, Scott Prevost, Matthew Stone.

## Animated Conversation:
## Modelling the Interaction between
## Gesture, Facial Expression, Intonation and Discourse

The Gesture-Jack group, co-supervised by Justine Cassell, Norm Badler and Mark Steedman, and in collaboration with animnl, works with *Jack* in the service of two goals: modelling the interaction between speech and gesture in order to test theories of their respective contributions to meaning, and the creation of realistic animated autonomous conversational agents.

### Modelling the Interaction between Speech and Gesture

In recent work it has been argued that spontaneous gesture produced unwittingly by speakers and the speech it accompanies form an integrated conceptual system [4]. Thus, gesture is not a *translation* of speech, or a passive irrelevant consequence of it. Gesture and speech are different communicative manifestations of one single mental representation. However, until now research on the relationship between gesture and speech has been difficult to evaluate because of its primarily descriptive character. One way to move from descriptive to predictive theories is via formal models, which point up gaps in knowledge and fuzziness in theoretical explanations.

We are building such a formal model of the gesture-speech relationship, embodied in a dialogue generation system comprising two autonomous but identical programs each driving an animated human figure, to generate simulated conversational interaction. Each agent is capable of generating speech, intonation and gesture as a function of the information structure of the discourse and the state of knowledge of the participants, causing the conversational agents to move their arms appropriately and in synchrony with the speech they utter. The dialogue generation program contains a formal, predictive and explanatory theory of the gesture-speech relationship (that gesture occurence and gesture types can be predicted on the basis of the informational and attentional status of discourse entities).

This animated conversation system allows us to test our theories of the gesture-speech relationship by displaying an instantiation of the theory via the animation of the conversational agents. For example, these animations can be shown to naive viewers to obtain their judgements as to the naturalness of the gestures, in order to assess the theory in comparison to others.

### Automatically Generating Conversational Behaviors in Animated Agents

One concern of research in computer graphics is to simulate human behavior in such a way that the animated agents truly display convincing and natural behavior. Of course, expert animators may be able to produce realistic cartoon figures either by depending on their intuitions about human behavior, or by capturing the motions of a human actor using real-time motion sensing devices. For the less computer-sophisticated user, or to free up the skilled animator to concentrate on other tasks, an alternative is to *generate* behavior on the basis of rules abstracted from the study of human behavior.

The behavior that we concentrate on in this project is conversation (that is, an interactive dialogue between two agents). Conversation includes spoken language (words and contextually appropriate intonation marking topic and focus), facial movements (lip shapes, emotions, gaze direction, head motion), and hand gestures (handshapes, points, beats, and motions representing the topic of accompanying speech). Without all of these verbal and non-verbal behaviors, one cannot have realistic, or at least believable, autonomous agents. To this end, our system *automatically* animates conversations between multiple human-like agents with appropriate and synchronized speech, intonation, facial expressions, and hand gestures.

## System Architecture

The same system is employed to realize both of the aforementioned theoretical goals.

- The *Dialogue Generation* program (see Matthew Stone) is a novel generalization of earlier work by  [5], and  [3] which allows gesture and conversational intonation to be generated along with speech. The output of the dialogue generation program is speech annotated with descriptions of appropriate intonation and gesture, which are then sent on to an intonation synthesis module and gesture synthesis module, respectively.

- The *Intonation Synthesis* model was designed by Scott Prevost and Mark Steedman and generates intonational tunes as a function of the information structure of the discourse.

- The *Facial Expression Specification* module, designed by Catherine Pelachaud, which generates head, eye and lip movements as a function of dialogic categories such as *planning what to say, controlling the listener's contribution to the conversation, feedback to speaker's contribution, comment on speaker's contribution*.

- The *Gesture Synthesis* module has two parts;

  - A *Synchronization* module, designed by Brett Douville. Interaction between agents and synchronization of gaze and hand movements to the dialogue for each agent are accomplished using Parallel Transition Networks (PaT-Nets), which allow coordination rules to be encoded as simultaneously executing finite state automata.

  - A *Movement Specification* module, built by Brett Achorn, which generates handshapes, and wrist and arm motion.

## Example

The conversation below is an example of the discourse output from the dialogue generation program. Following it is a description of some of the nonverbal and intonational behaviors generated with the speech.

The dialogue is unnaturally repetitive and explicit in its goals because the dialogue generation program that produced it has none of the conversational inferences that allow humans to follow leaps of reasoning. Therefore, the two agents have to specify in advance each of the goals they are working towards and steps they are following.

```
Gilbert:      Do you have a blank check?
George:       Yes, I have a blank check.
Gilbert:      Do you have an account for the check?
George:       Yes, I have an account for the check.
Gilbert:      Does the account contain at least fifty dollars?
George:       Yes, the account contains eighty dollars.
Gilbert:      Get the check made out to you for fifty dollars\
              and then I can withdraw fifty dollars for you.
George:       All right, let's get the check made out to me
              for fifty dollars.
```

When Gilbert asks a question, his voice rises. When George replies to a question, his voice falls. When Gilbert asks George whether he has a blank check, he stresses the word "check". When he asks George whether he has an account for the check, he stresses the word "account".

Every time Gilbert replies affirmatively ("yes"), or turns the floor over to Gilbert ("all right"), he nods his head, and raises his eyebrows. George and Gilbert look at each other when Gilbert asks a question, but at the end of each question, Gilbert looks up slightly. During the brief pause at the end of affirmative statements the speaker (always George, in this fragment) blinks. To mark the end of the questions, Gilbert raises his eyebrows.

In saying the word "check", Gilbert sketches the outlines of a check in the air between him and his listener. In saying "account", Gilbert forms a kind of box in front of him with his hands: a metaphorical representation of a bank account in which one keeps money. When he says the phrase "withdraw fifty dollars," Gilbert withdraws his hand towards his chest.

## Publications

Two papers describe our research: [1], and [2].

# Information Theory Reading Group

Department of Computer and Information Science

Robin Clark, Jason Eisner, Aravind Joshi, Mitch Marcus, Dan Melamed, Lance Ramshaw, Jeff Reynar, Joseph Rosenzweig, B. Srinivas, Matthew Stone, David Yarowsky.

## Information Theory Reading Group

This group has met weekly throughout the year, working through the Cover & Thomas text *Elements of Information Theory*. Areas of particular focus have included Markov models, text coding and compression, gambling and the stock market, Kolmogorov complexity theory, and the relationship between information theory and statistics.

# The STAG Machine Translation Project

Department of Computer and Information Science

Hyun Seok Park, Dania Egedi, Martha Palmer, Aravind K. Joshi.

## Introduction

It is well-understood that accurate machine translation often requires reference to contextual knowledge for the correct treatment of linguistic phenomena such as pronoun reference and gender agreement [3]. This is still, in many cases, an unsolved problem for natural language analysis [5], which adds to the burden of the already beleaguered machine translation systems. One of the historical arguments in favor of the interlingua approach has been that, since it revolves around a deep semantic representation, it is better able to handle pronoun reference and other linguistic phenomena that are seen as requiring a knowledge-based approach. However, recent implementations of machine translation systems are blurring the distinction between transfer systems and interlingua systems. Wilks [9] discusses the hybrid IBM system which achieves improved coverage by combining certain transfer based techniques with what was fundamentally a statistical approach. Nirenburg [4] is currently proposing a multi-engine system which will apply more than one approach to a sentence and then choose the best result. The STAG Machine Translation Project is developing a prototype system for machine translation between English and Korean. The system is implemented in the Synchronous TAG formalism [7], which is an extension of lexicalized [6], feature-based [8] Tree Adjoining Grammars [2] (FB-LTAGs). Although this is essentially a transfer based approach, it uses feature unification for lexical selection and is being augmented with a discourse model to handle discourse related phenomena such as recovery of topicalized arguments.

Our domain consists of a set of English military messages that have been translated into Korean by three native speakers. The messages are in the form of transcripts of messages sent over short-wave radio, and comprise short dialogues between two individuals. The example given below is a paraphrase of an existing message:

```
[000001 Subject:BATTLE RESOURCES - 2 REPEATS]
REQUEST BREAKDOWN OF BATTLE RESOURCES.  WHAT HAPPENED TO THE XX-10S?
YOUR LAST MESSAGE  SHOWED NONE ON HAND.  NEED COMMANDERS REPORT.
```

The eventual goal of this project is to translate these messages bi-directionally from English to Korean and from Korean to English. Although neither direction is trivial, translation from Korean to English is the more difficult direction. The structure of Korean allows information to be inferred from previous mention or context, while English usually requires that the information be explicit in each sentence. This manifests itself in a lack of definite and indefinite determiners and dropped noun phrases in Korean which must be recovered before the sentence can be translated into English.

Recent work on the project [1] includes a developing a small Korean grammar in TAGs that handles linguistic phenomena such as simple relative clauses and wh-questions, as well as the mappings needing to translate such structures between Korean and English; work on lexical selection in translating from Korean to English; and recovering dropped NPs in Korean in preparation for translating to English.

Please see notes by Martha Palmer and Hyun Seok Park for more information on work being done on machine translation, as well as the XTAG project for information on more general work being done in TAGs.

# The TraumAID Project

Department of Computer and Information Science

Bonnie Lynn Webber, John R. Clarke, M.D., Diane Chi, Abigail Gertner, Jonathan Kaye, Stefanie Neumann, Lola Ogunyemi, Moninder Singh.

Injury is a major health problem in the United States, resulting in more years of human life lost than any other disease. Believing that the morbidity and mortality due to injury can be reduced through rapid delivery of expert care, the American College of Surgeons developed an Advanced Trauma Life Support (ATLS) course that educates physicians in the *first phase* of trauma management: the initial evaluation, resuscitation, and stabilization of severely injured patients. For this same reason, we have been developing TRAUMAID to assist physicians in the *second phase*: the initial definitive management of those patients in Emergency Centers and Trauma Centers.

During this second phase, a medical team led by the attending physician acts to identify the patient's injuries, providing initial therapy and preparing for further diagnosis and therapy in the X-Ray Department, Operating Room (OR) or Intensive Care Unit (ICU). This phase, which can last up to two hours, is often characterized by the need for urgent actions and definitive decisions that preempt further diagnostic (i.e., information-gathering) activity: e.g., the patient must be taken to the OR immediately and pending goals satisfied through surgical procedures. Decision support during this second phase of trauma management must therefore balance systematic diagnostic activity with the demands of quick and definitive therapeutic action.

Work has been proceeding on TraumAID for the past ten years, involving members of the Medical College of Pennsylvania (MCP) led by Dr. John Clarke (Professor of Surgery), and students and faculty at the University of Pennsylvania, led by Professor Bonnie Webber. Recently, Professor Sandra Carberry, of the University of Delaware, has joined the group, to investigate issues of user modelling in the context of information delivery. The current working system *TraumAID 2.0* comprises (1) a *rule-based reasoner* that addresses the question: given the current situation (i.e., what is known about the patient), what *conclusions* can be drawn and therefore what *goals* are most appropriate for the physician to adopt; and (2) a *planner* that addresses the question: given the entire set of currently relevant goals, what *actions* should the physician and medical team best perform now.

This division into local reasoner and global planner reflects the demands noted above: first, it supports *goal-directed diagnosis*, in the sense that its attempt to characterize a patient's injuries is not pursued beyond the point that it would make a difference to the therapeutic goals that one would adopt under the circumstances. Second, it engenders flexibility in patient management. As more is learned about the patient's condition through diagnostic activity, TraumAID's advice on what to do next will constantly reflect the entire set of current management goals. The physician can thus be directed to actions that can be used to satisfy multiple goals and to actions that are of greatest urgency.

TraumAID 2.0 currently runs on SUN workstations and high-end versions of the MAC, both using a menu-based interface for data entry. One important issue of concern to us is clinically acceptable and effective forms of *information delivery*, with Abigail Gertner developing a method based on *reactive critiquing* in non-urgent situations and *proactive advising* in urgent situations. (See CLiFF Note for Abigail Gertner.) We are also experimenting with the use of *automatic speech* *generation* to deliver both critique and advice, augmenting the textual record that appears on the computer screen. (See CLiFF Note for Scott Prevost.)

Other current development work on TraumAID is proceeding in three directions. The first involves development of a computer model of acute cardio-vascular response that can be used to interpret changes in a patient's vital signs in response to blood loss and fluid replacement and thus further aid in diagnosis. Since we are constructing a general model, when complete, it should be of use in other systems than TraumAID – e.g., for modeling gastro-intestinal or interoperative bleeding. This work is being done by Stefanie Neumann, a PhD student in Bioengineering.

Second, we are exploring the use of 3-D human figure modeling techniques pioneered by Penn's Center for Human Modelling and Simulation, both to augment a physician's own anatomical visualization abilities and to improve TraumAID's ability to reason the relationship between anatomical structure and physiological function and its disturbance through injury. (See CLiFF Note for Jonathan Kaye.)

Third, in another joint project with Penn's Center for Human Modelling and Simulation, we are developing a system (MediSim) to provide a virtual environment in which medical personnel (either human avatars or simulated humans) can interact with simulated trauma patients for purposes of training and/or evaluation. Work on MediSim began in Fall 1994, supported by ARPA's Advanced Biomedical Technology initiative. Current work involves both support of realistic patient models (work being done by CIS graduate student, Diane Chi) and development of a larger repertoire of diagnostic and therapeutic actions.

# The XTAG Project

Department of Computer and Information Science

Christy Doran, Dania Egedi, Beth Ann Hockey, B. Srinivas, Tilman Becker, Seth Kulick, Anoop Sarkar, Aravind K. Joshi.

## Introduction

XTAG [1] is an on-going project to develop a wide-coverage grammar for English, based on the Feature-Based Lexicalized Tree Adjoining Grammar (FB-LTAG) formalism. FB-LTAG is a lexicalized [5] mildly-context sensitive tree rewriting system [3] that is closely related to Dependency Grammars and Categorial Grammars. XTAG also serves as an FB-LTAG development system consisting of a predictive left-to-right parser, an X-window interface, a morphological analyzer, and a part-of-speech tagger.

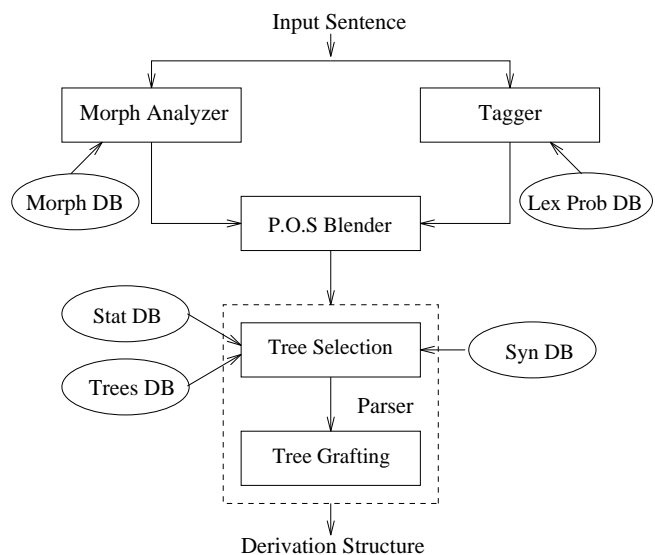

Figure 1: Overview of the XTAG system

## System Description

Figure 1 shows the overall flow of the system when parsing a sentence. The input sentence is submitted to the **Morphological Analyzer** and the **Tagger**. The morphological analyzer retrieves the morphological information for each word from the morphological database. This output is filtered in the **P.O.S Blender** using the output of the trigram tagger to reduce the part-of-speech ambiguity of the words. The sentence, now annotated with part-of-speech tags and morphological information for each word, is input to the **Parser**, which consults the syntactic database and tree database to retrieve the appropriate tree structures for each lexical item. A variety of heuristics are used to reduce the number of trees selected. The parser then composes the structures to obtain the parse(s) of the sentence.

The details of each component are summarized below:

**Morphological Analyzer and Morph Database**
- Consists of approximately 317,000 inflected items.
- Entries are indexed on the inflected form and return the root form, POS, and inflectional information.
- Database does not address derivational morphology.

**POS Tagger and Lexical Probability Database**
- Wall Street Journal-trained trigram tagger extended to output N-best POS sequences.
- Decreases the time to parse a sentence by an average of 93%.

**Syntactic Database**
- 37,000 entries.
- Each entry consists of: the uninflected form of the word, its POS, the list of trees or tree-families associated with the word, and a list of feature equations that capture lexical idiosyncrasies.

**Tree Database**
- 385 trees, divided into 40 tree families and 62 individual trees.
- Tree families represent subcategorization frames; the trees in a tree family would be related to each other transformationally in a movement-based approach.

**X-Interface**
- Menu-based facility for creating and modifying tree files
- User controlled parser parameters: parser's start category, enable/disable/retry on failure for POS tagger.
- Storage/retrieval facilities for elementary and parsed trees as text and postscript files.
- Graphical displays of tree and feature data structures.
- Hand combination of trees by adjunction or substitution for diagnosing grammar problems.

### English Grammar

The morphological, syntactic, and tree databases together comprise the English grammar. Lexical items not in the databases are handled by default mechanisms. The range of syntactic phenomena that can be handled is large and includes auxiliaries (including inversion), copula, raising and small clause constructions, topicalization, relative clauses, infinitives, gerunds, passives, adjuncts, it-clefts, wh-clefts, PRO constructions, noun-noun modifications, extraposition, determiner phrases, genitives, negation, noun-verb contractions and imperatives. The combination of a large lexicon and wide phenomena coverage result in a robust system. The XTAG grammar has been relatively stable since November, 1993, although new analyses are still being added periodically. Analyses of time NPs and sentential adjuncts are currently under development.

### Recent Developments

**Improved Part-of-speech Tagging**: We have incorporated a trigram part-of-speech tagger [4] trained on the Wall Street Journal Corpus into the XTAG system. The system had previously included a tagger trained on the Brown Corpus tags, but the current tagger has been trained on the XTAG tags (a subset of the Brown tags); this eliminates the need to translate the tags, and decreases the ambiguity of the tags output by the tagger. In addition, the tagger has been extended to output the N-best parts-of-speech sequences [6]. XTAG uses this information to reduce the number of specious parses by filtering the possible parts-of-speech provided by the morphological analyzer for each word. The tagger decreases the time to parse a sentence by an average of 93%.

**Implementation of Simulated Multicomponent Adjunction**: The mechanism for simulated multicomponent adjunction developed by Hockey and Srinivas [2] has been used in the implementation of auxiliary inversion. Simulated multicomponent adjunction is also being implemented to handle extraposition and extraction from adjuncts.

See notes in this volume by B. Srinivas, Dania Egedi, Christy Doran, Tilman Becker, Seth Kulick, and the STAG Machine Translation Project for more information on work related to TAGs.

XTAG is available via ftp. Instructions and more information can be obtained by mailing requests to xtag-request@linc.cis.upenn.edu.

**Part IV**

# Appendix

# CLiFF Talks

## Spring 1994

Feb 1     Bernhard Rohrbacher and Tom Roeper     UPenn—IRCS and UMass
Economy of Projection and Pro-Drop in Child English

Feb 8     Lisa Rau                                UPenn—CIS
Calculating Salience of Knowledge

Feb 10     Mark Dras                              Macquarie U and Microsoft
PP Attachment Using Role Information

Feb 22     Lance Ramshaw                          UPenn—IRCS
Exploring the Behavior of "Brill Rules"

Mar 15     Libby Levison                          UPenn—CIS
How Animated Agents Perform Tasks

Mar 22     Michael White                          UPenn—-CIS
A Sortal Approach to Aspectual Composition

Mar 25     Daniela Rus                            Cornell
Finding Information Requires More Than Words

Mar 28     Igor Boguslavsky                       Russian Academy of Sciences
Computer Implementation of the Meaning/Text Theory

Mar 29     Mark Johnson                           Brown
Featurs and Formulae

Apr 19     Dania M. Egedi                         UPenn—CIS
Learning Through Metaphor

May 9     Michael Niv                             Technion
A Psycholinguistically Motivated Parser for CCG

May 10     Francesc Ribas                         Upenn—CIS
Some Experiments on Learning Appropriate Verbal Selectional Restrictions

May 18     Jeffrey Siskind                        Toronto
Spanning Intervals for Temporal Inference and Event Perception

## Fall 1994

Sep 9    Catherine Pelachaud and Scott Prevost    UPenn—CIS
Sight and Sound: Generating Facial Expressions and Spoken Intonation

Sep 20    Uzzi Ornan and Michael Katz    Technion
Natural Language Processing of Hebrew

Sep 26    Jason Eisner    UPenn—CIS
'A'-less in Wonderland: Revisiting 'Any'

Sep 27    Abigail Gertner    UPenn—CIS
Upholding the Maxim of Relevance during Patient-Centered Activities

Oct 10    Thad Polk and Martha Farah    UPenn—Psych
A Simple Common Contexts Explanation for the Development of Abstract Letter Identities

Oct 11    Carl Gunter    UPenn—CIS
The Common Order-theoretic Structure of Version Spaces and ATMS's

Nov 1    James Rogers    UPenn—CIS
On Descriptive Complexity, Language Complexity and GB

Nov 8    Stefano Bertolo    Rutgers
General Constraints on Maturational Solutions for Triggering Learning

Nov 15    Spyridoula Varlokosta    UPenn—IRCS
Factive Islands

# 1994 LINC Lab Technical Reports

**Massively Parallel Simulation of**
**Structured Connectionist Networks: An Interim Report**
*D.R. Mani*
*Lokendra Shastri*
**MS-CIS-94-10**
**LINC LAB 264**

**SodaJack: An Architecture for Agents**
**That Search For and Manipulate Objects**
*Christopher Geib*
*Libby Levison*
*Micheal Moore*
**MS-CIS-94-16**
**LINC LAB 265**

**The Well-tempered Computer**
*Mark Steedman*
**MS-CIS-94-19 LINC LAB 266**

**Process Algebra, CCS, and Bisimulation**
**Decidability**
*Seth Kulick*
**MS-CIS-94-20**
**LINC LAB 267**

**Modeling the Interaction between**
**Speech and Gesture**
*Justine Cassell*
*Matthew Stone*
*Brett Douville*
*Scott Prevost*
*Brett Achorn*
*Mark Steedman*
*Norm Badler*
*Catherine Pelachaud*
**MS-CIS-94-23**
**LINC LAB 268**
**HUMAN MODELING & SIMULATION LAB 61**

**Centering: A Framework for Modelling the Coherence of Discourse**
*Barbara J. Grosz*
*Aravid K. Joshi*
*Scott Weinstein*
**MS-CIS-94-40**
**LINC LAB 274**

**Description Based Parsing In A Connectionist Network (Dissertation)**
*James B. Henderson*
**MS-CIS-94-46**
**LINC LAB 275**

**Binding and Control In CCG and Its Relatives**
*Mark Steedman*
**MS-CIS-94-47**
**LINC LAB 276**

**A Multiple-Conclusion Meta-Logic**
*Dale Miller*
**MS-CIS-94-51**
**LINC LAB 277**

**Formal and Computational Aspects of Natural Language Syntax (Dissertation)**
*Owen Rambow*
**MS-CIS-94-52**
**LINC LAB 278**

**A Computational Approach to Aspectual Composition (Dissertation)**
*Michael White*
**MS-CIS-94-53**
**LINC LAB 279**

**Planning and Terrain Reasoning**
*Michael B. Moore*
*Christopher Geib*
*Barry D. Reich*
**MS-CIS-94-63**
**LINC LAB 280**

**Korean Grammar Using Tags (Masters Thesis)**
*Hyun Seok Park*
**MS-CIS-94-64**
**LINC LAB 281**